\documentclass{ejasa}
\usepackage[english]{babel}
\usepackage{natbib}
\usepackage[english]{babel}
\usepackage[utf8x]{inputenc}
\usepackage[T1]{fontenc}
\usepackage{dcolumn}
\usepackage[a4paper,top=3cm,bottom=2cm,left=3cm,right=3cm,marginparwidth=1.75cm]{geometry}

\usepackage{amsmath}
\usepackage{graphicx}
\usepackage[colorinlistoftodos]{todonotes}

\usepackage{lineno}
\usepackage{subfigure}
\usepackage{todonotes} 
\usepackage{tabularx}
\usepackage{multirow}
\usepackage{hhline}
\usepackage{comment}
\title{Blaming humans in autonomous vehicle accidents: Shared responsibility across levels of automation}

\newcommand{\authorlastnames}{Awad, Levine et al.}

\author[a,+]{Edmond Awad}
\author[a,b,c,+]{Sydney Levine} 
\author[b]{Max Kleiman-Weiner}
\author[a]{Sohan Dsouza}
\author[b,*]{Joshua B. Tenenbaum}
\author[d,*]{Azim Shariff}
\author[e,*]{Jean-Fran\c{c}ois Bonnefon}
\author[a,f,*]{Iyad Rahwan}

\affil[a]{Media Lab, Massachusetts Institute of Technology, MA, USA}
\affil[b]{Department of Brain and Cognitive Sciences, Massachusetts Institute of Technology, MA, USA}
\affil[c]{Department of Psychology, Harvard University, MA, USA}
\affil[d]{Department of Psychology and Social Behavior, University of California, Irvine, CA, USA}
\affil[e]{Toulouse School of Economics (TSM-R), CNRS, University of Toulouse, France}
\affil[f]{Institute for Data, Systems and Society, Massachusetts Institute of Technology, MA, USA}
\affil[+]{Joint first author}
\affil[*]{Corresponding authors. e-mail: jbt@mit.edu; shariffa@uci.edu; jean-francois.bonnefon@tse-fr.eu; irahwan@mit.edu}

\date{}

\begin{document}
\maketitle
%
%




\begin{abstract}
\textbf{
When a semi-autonomous car crashes and harms someone, how are blame and causal responsibility distributed across the human and machine drivers? In this article, we consider cases in which a pedestrian was hit and killed by a car being operated under shared control of a primary and a secondary driver. We find that when only one driver makes an error, that driver receives the blame and is considered causally responsible for the harm, regardless of whether that driver is a machine or a human. However, when both drivers make errors in cases of shared control between a human and a machine, the blame and responsibility attributed to the machine is reduced. This finding portends a public under-reaction to the malfunctioning AI components of semi-autonomous cars and therefore has a direct policy implication: a bottom-up regulatory scheme (which operates through tort law that is adjudicated through the jury system) could fail to  properly regulate the safety of shared-control vehicles; instead, a top-down scheme (enacted through federal laws) may be called for.
}
\end{abstract}

\section{Introduction}
Every year, about 1.25 million people die worldwide in car crashes \citep{who2017injuries}. Laws concerning principles of negligence currently adjudicate how responsibility and blame get assigned to the individuals who injure others in these harmful crashes. The impending transition to fully autonomous cars promises a radical shift in how blame and responsibility will get attributed in the cases where crashes do occur, but most agree that little or no blame will be attributed to the occupants in the car, who will, by then, be entirely removed from the decision-making loop \citep{Geistfeld2017roadmap}. However, before this era of fully autonomous cars arrives, we are entering a delicate era of shared control between humans and machines. 

This new moment signals a departure from our current system -- where individuals have full control over their vehicles and thereby bear full responsibility for crashes -- to a new system where blame and responsibility may be shared between a human and a machine driver. The spontaneous reactions people have to crashes that occur when a human and machine share control of a vehicle has at least two direct industry-shaping implications.  First, at present, little is known about how the public is likely to respond to crashes that involve both human and machine drivers.  This uncertainty has concrete implications: manufacturers price products to reflect the liability they expect to incur from the sale of those products. If manufacturers cannot assess the scope of the liability they will incur from semi-autonomous and autonomous vehicles, that uncertainty will translate to substantially inflated prices of AVs. Moreover, the rate of the adoption of semi-autonomous and autonomous vehicles will be proportional to the cost to consumers to adopt the new technology \citep{Geistfeld2017roadmap}.  Accordingly, the uncertainty about the extent of corporate liability for semi-autonomous and autonomous vehicle crashes is quantitatively slowing down AV adoption, while millions of people continue to die in car crashes each year.  Clarifying how and when responsibility will be attributed to manufacturers in semi-autonomous and autonomous car crashes will be a first step in reducing this uncertainty and speeding the adoption of semi-autonomous and eventually autonomous vehicles.  

The second direct implication of this work will be to forecast how a tort-based regulatory scheme (which is decided on the basis on jury decisions) is likely to turn out.  Put another way, understanding how the public is likely to react to crashes that involve both a human and a machine driver will give us a hint to how the laws will be shaped if we let jury decisions create the regulations. If our works uncovers systematic biases that are likely to impact juries and would impede the adoption of semi-autonomous and autonomous cars, then it may make sense for federal regulations be put in place, which would preempt the tort system from regulating these cars.

Already, semi-autonomous vehicle crashes are in the public eye.  In May 2016, the first deadly crash of a Tesla Autopilot car occurred and the occupant of the car was killed. In a news release, Tesla explained: ``Neither Autopilot nor the driver noticed the white side of the tractor-trailer against a brightly lit sky, so the brake was not applied'' \citep{tesla2016loss}. That is, both the machine driver and the human driver should have taken action (breaking to avoid a truck making a left turn across traffic) and neither driver did. The mistakes of both drivers led to the crash. The National Highway Safety Traffic Administration (NHSTA) did an investigation of the incident and did not find Tesla at fault in the crash \citep{nhtsa2016investigation}, and a statement from the family of the killed occupant indicated that they did not hold the car responsible for what happened \citep{reuters2017family}. However, press attention surrounding the incident was markedly skewed towards blaming the human driver for the crash, with rumors quickly circulating that the driver had been watching a Harry Potter movie (eg, \citep{ap2016harrypotter}), though upon further investigation it was discovered that there was actually no evidence grounding this claim \citep{chong2017notharrypotter}. This set of anecdotes around the Tesla crash begins to suggest a troubling pattern, namely, that humans might be blamed more than their machine partners in certain kinds of semi-autonomous vehicle crashes. Was this pattern a fluke of the circumstances of the crash and the press environment?  Or does it reflect something psychologically deeper that may color our responses to human-machine joint action, and in particular, when a human-machine pair jointly controls a vehicle?

\begin{figure}[!ht]
\centering
\includegraphics[width=1.0\linewidth]{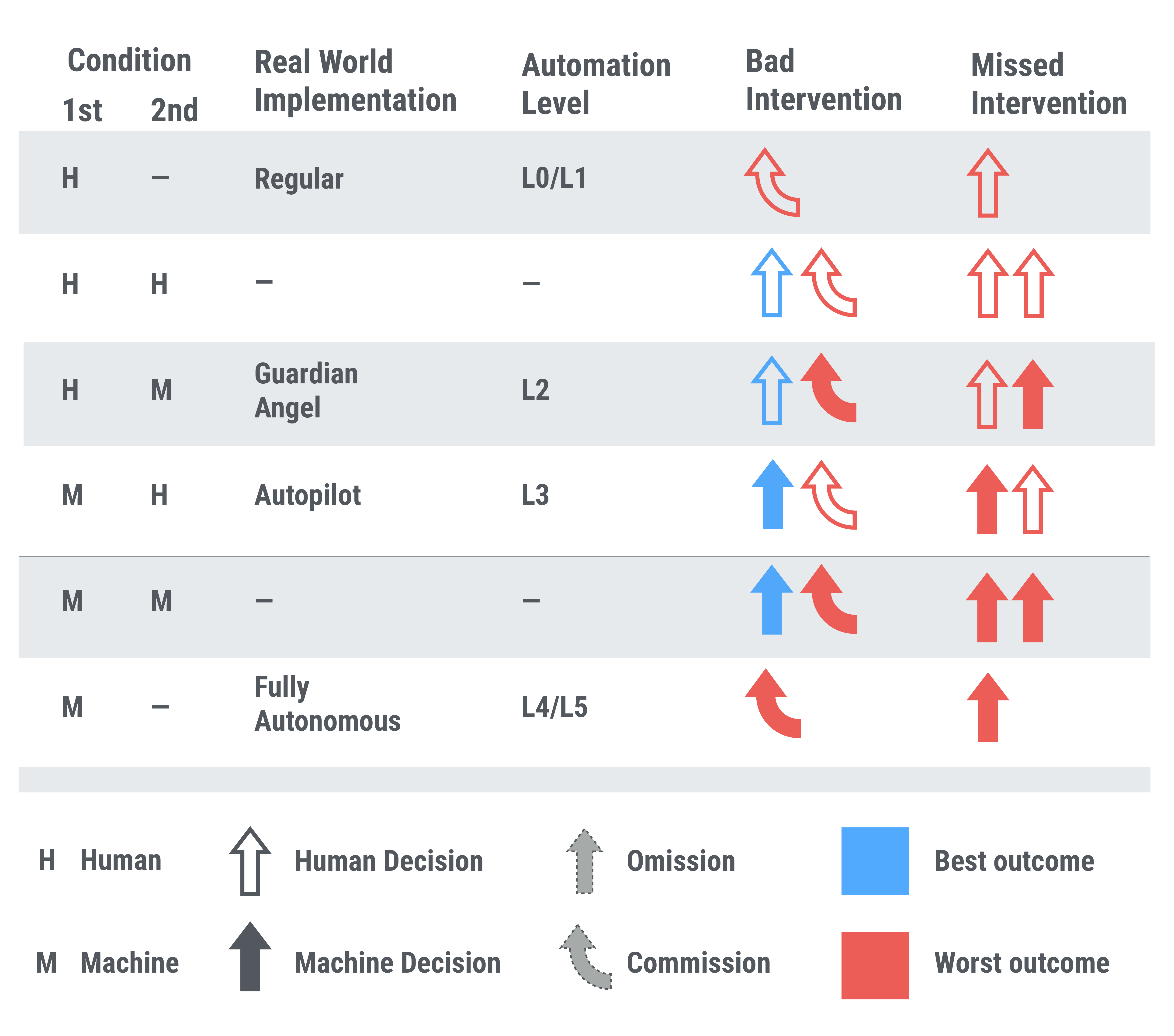}
\caption{\textbf{Actions or action sequences for the different car types considered.} Outline arrows indicated an action by a human `H', and solid arrows indicate an action by a machine `M'. The top and bottom rows represent single-driver cars, while all others represent dual-driver cars. A red arrow indicates a decision -- whether action or inaction -- with the avoidable death of a pedestrian as the outcome, and a blue arrow indicates a decision that does not result in any deaths. For example, the H+M type, which corresponds to NHTSA Level 2 and is implemented as the Guardian Angel system, has a human main driver and a machine standby driver. A Bad Intervention then involves the machine killing the pedestrian by overriding (solid, angled, red arrow) the human's non-lethal staying of course (outline, straight, blue arrow). A Missed Intervention involves the human staying on course to kill the pedestrian (outline, straight, red arrow) without intervention from the machine (solid, straight, red arrow).}
\label{fig:design}
\end{figure}

What we are currently witnessing is a gradual increase toward full automation, going through several steps of shared control between user and vehicle, which may take decades due to technical and regulatory issues as well as attitudes of consumers towards adoption \citep{munster2017normal,Kessler2017timeline} (see Figure \ref{fig:design}). Some vehicles can take control over the actions of a human driver (e.g., Toyota's `Guardian Angel') to perform emergency maneuvers. Other vehicles may do most of the driving, while requiring the user to constantly monitor the situation and be ready to take control (e.g., Tesla's `Autopilot').

Our central question is this: when a semi-autonomous car crashes and harms someone, how are blame and causal responsibility distributed across the human and the machine drivers? In this article, we consider cases in which a pedestrian was hit and killed by a car being operated under shared control of a primary and a secondary driver. We consider a large range of control regimes (see Figure \ref{fig:design}), but the two main cases of interest are the instances of shared control where a human is the primary driver and the machine a secondary driver (``human-machine'') and where the machine is the primary driver and the human the secondary driver (``machine-human''). We consider a simplified space of scenarios in which (a) the main driver makes the correct choice and the secondary driver incorrectly intervenes (``Bad Intervention'') and (b) the main driver makes an error and the secondary driver fails to intervene (``Missed Intervention'').  Both scenarios end in a crash.  For comparison, we also include analogous scenarios involving a single human driver (a regular car) or a single machine driver (a full autonomous car) as well as two hypothetical two-agent cars (driven by two humans or two machines).

In Bad Intervention cases, the primary driver (be it human or machine) has made a correct decision to keep the car on course, which will avoid a pedestrian. Following this, the secondary driver makes the decision to swerve the car into the pedestrian. In these sorts of cases, we expect that the secondary driver (the only driver that makes a mistake) will be blamed more than the first driver. What is less clear is if people will assign blame and causal responsibility differently if this secondary driver is a human driver or a machine. Recent research suggests that robots may be blamed more than humans for making the same decisions in the same situations \citep{malle2015sacrifice}. Likewise, people tend to lose trust in an algorithm that makes an error faster than they lose trust in a human that makes the same error \citep{dietvorst2015algorithm}.  Together, this research suggests that the machine drivers may be judged more harshly overall than human drivers who make the same mistakes.

In Missed Intervention cases, the primary driver has made an incorrect decision to keep the car on course (rather than swerving), which would cause the car to hit and kill a pedestrian. The secondary driver then neglects to swerve out of the way of the pedestrian. In these cases, the predictions for how subjects will distribute blame and causal responsibility are less clear because both drivers make a mistake. As in the Bad Intervention cases, agent type (human or machine) may have an effect on blame and causal responsibility ratings.  But unlike with Bad Intervention cases, Missed Intervention cases introduce the possibility that driver role (primary or secondary) may also impact judgments. It is possible that subjects may shift responsibility and blame either toward the agent who contributed the most to the outcome (primary driver), or to the agent who had the last opportunity to act (secondary driver; \citep{chockler2004responsibility,gerstenberg2012contributions,sloman2015causality,zultan2012finding}. Under some regimes -- such as Toyota's Guardian Angel -- the user does most of the driving, but the decision to override (and thus to act last) pertains to the machine. Under others -- such as Tesla's autopilot -- the machine does most of the driving, but the decision to override pertains to the user. 

\begin{figure*}
\centering
\includegraphics[width=0.99\linewidth]{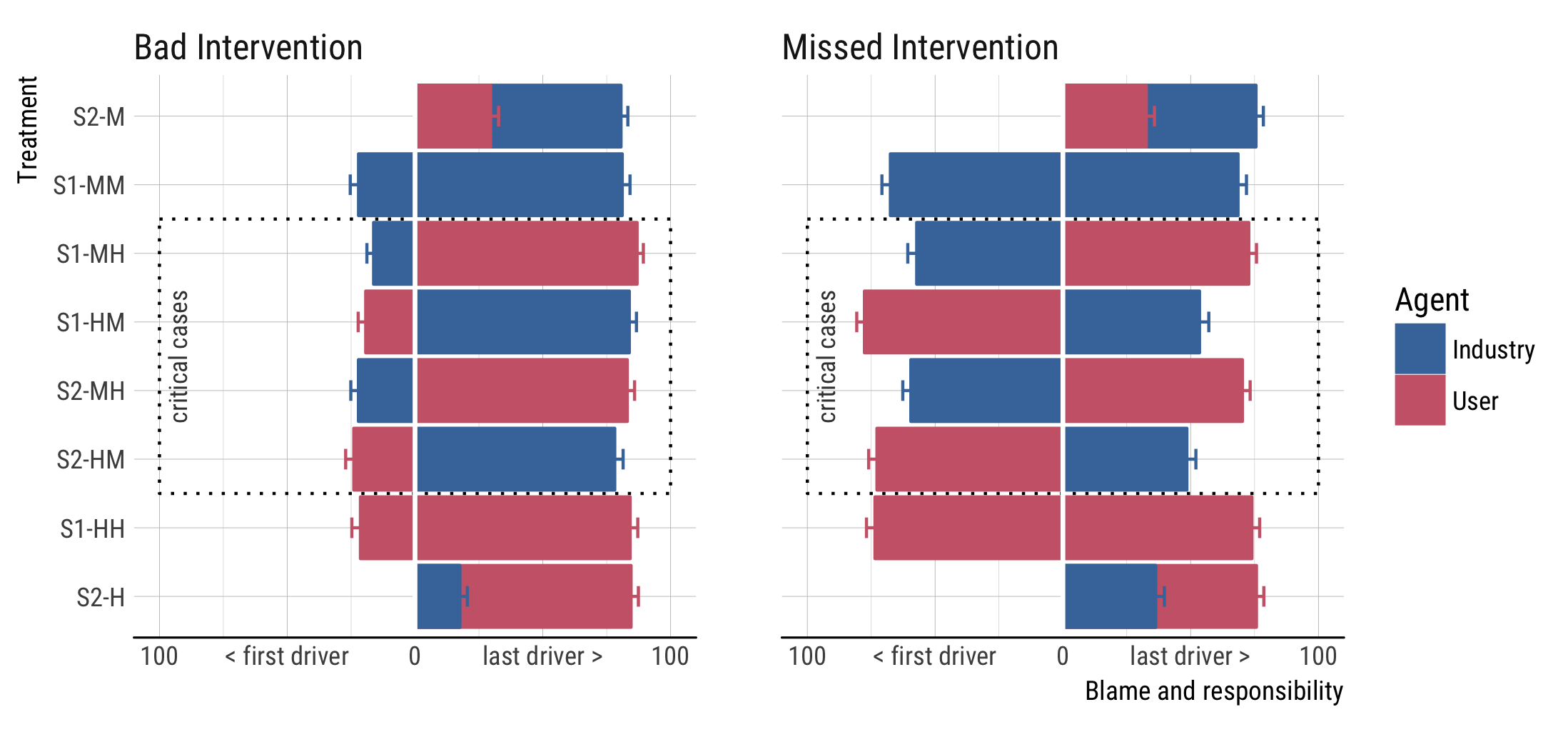}
\caption{\textbf{Results from studies 1 and 2 for six car types (y-axis) in two cases; Bad Intervention (left) and Missed Intervention (right).} Ratings of blame and causal responsibility  are aggregated (collectively referred to as blame, henceforth). For Industry, when applicable, ratings of car and company are aggregated (collectively referred to as Industry, henceforth). The y-axis represents the six car types considered in the two studies, S1 and S2. Two car types, HM (human-machine) and MH (machine-human), were considered in both studies. The y-axis labels include the study and the car type. For example, S1-HM represents the Human-Machine regime ratings collected in study 1. In the six car types, the x-axis labeling of first driver refers to the main driver, and the last driver refers to the secondary driver in dual-agent cars, and the sole driver in the single-agent cars. Industry and User ratings are shown in blue, and red, respectively. For Bad Intervention, only one agent has erred (the last driver). This agent (whether User or Industry) is blamed almost the same across all car types, and is always blamed more than the other agent (first driver) in every type (row). For Missed Intervention, in dual-agent cars (rows 2-7), both agents have erred (by avoiding intervention). When both agents are the same, that is both are humans or both are machines (rows 2 and 7), both are blamed similarly. However, when human and machine are sharing control (critical cases, inside the dotted rectangle), blame ratings of Industry drops significantly, regardless of the role of the machine (main or secondary). This blame drop is not mirrored in the Bad Intervention case, when only the machine has erred.}
\label{fig:mainresults}
\end{figure*}

In both Bad and Missed Intervention cases, we also investigate whether blame and causal responsibility judgments can be explained by judgments of the competence of the drivers.  People sometimes give an agent less blame for an outcome when they believe that the agent lacked the ability to perform that action reliably \citep{knobe2003intentional}.  If subjects think that the machine driver in a semi-autonomous car is less competent than the human driver, they may be less likely overall to assign blame to the machine (although it is also possible that subjects may find a manufacturer \textit{more} blameworthy for deploying an unreliable vehicle).

\section{Results}

\begin{table*}[!ht] \centering \tiny
\begin{tabular}{@{\extracolsep{5pt}}lcccccc} 
\\[-1.8ex]\hline 
\hline \\[-1.8ex] 
 & \multicolumn{6}{c}{Blame and Causal Responsibility} \\ 
\cline{2-7} 
 & \multicolumn{3}{c}{Bad Intervention} & \multicolumn{3}{c}{Missed Intervention} \\ 
\\[-1.8ex] & \multicolumn{1}{c}{Study 1} & \multicolumn{1}{c}{Study 2} & \multicolumn{1}{c}{Study 3} & \multicolumn{1}{c}{Study 1} & \multicolumn{1}{c}{Study 2} & \multicolumn{1}{c}{Study 3}\\ 
\hline \\[-1.8ex] \\
 Human & 1.192 & 1.810 & -1.652 & 17.321$^{***}$ & 13.268$^{***}$ & 2.341 \\ 
  & (1.500) & (2.226) & (0.939) & (1.628) & (2.526) & (0.959) \\ 
  & & & & & & \\ 
 Mistake & 63.347$^{***}$ & 56.011$^{***}$ & 11.645$^{***}$ &  &  &  \\ 
  & (1.497) & (2.226) & (0.939) &  &  &  \\ 
  & & & & & & \\ 
 Human:Mistake & 1.899 & 3.096 & -0.092 &  &  &  \\ 
  & (2.121) & (3.148) & (1.329) &  &  &  \\ 
  & & & & & & \\ 
 Last Driver &  &  &  & -1.444 & -10.985$^{***}$ & -0.244 \\ 
  &  &  &  & (1.470) & (2.526) & (0.959) \\ 
  & & & & & & \\ 
 Human:Last Driver &  &  &  & -0.759 & 8.451 & 2.782 \\ 
  &  &  &  & (2.309) & (4.045) & (1.356) \\ 
  & & & & & & \\ 
 Constant & 19.124$^{***}$ & 22.138$^{***}$ & 28.089$^{***}$ & 60.316$^{***}$ & 59.500$^{***}$ & 36.740$^{***}$ \\ 
  & (1.057) & (1.586) & (1.015) & (1.634) & (2.138) & (1.083) \\ 
  & & & & & & \\ 
\hline \\[-1.8ex] 
Subject Random Effects? & Yes & Yes & Yes & Yes & Yes & Yes \\ 
Question Random Effects? & Yes & Yes & Yes & Yes & Yes & Yes \\ 
N & \multicolumn{1}{c}{786} & \multicolumn{1}{c}{382} & \multicolumn{1}{c}{586} & \multicolumn{1}{c}{786} & \multicolumn{1}{c}{382} & \multicolumn{1}{c}{586} \\ 
Observations & \multicolumn{1}{c}{3,144} & \multicolumn{1}{c}{1,528} & \multicolumn{1}{c}{4,688} & \multicolumn{1}{c}{3,144} & \multicolumn{1}{c}{1,528} & \multicolumn{1}{c}{4,688} \\ 
\hline 
\hline \\[-1.8ex] 
\multicolumn{7}{l}{$***$,$**$,$*$ significant at $0.01\%$, $0.1\%$ and $1\%$ levels, respectively.} \\ 
\end{tabular} 
 \caption{\textbf{Regression analysis of data collected in studies 1, 2, and 3 in the cases of Bad Intervention and Missed Intervention.} {\normalfont Studies 2 and 3 are limited to shared-control regimes. ``Human'' refers to the type of agent in question (that is, human as compared to the baseline, machine), ``Mistake'' refers to whether the decision was a mistake (that is, the decision would have resulted in losing a life, or losing more lives in study 3), and ``Last Driver'' refers to the driver role (that is, the driver assumes the secondary role). All models include subject random effects and question (blame or causal responsibility) random effects.}} 
  \label{tab:regression} 
\end{table*}

Study 1 compared four kinds of cars with different regimes of control. Each car had a primary driver, whose job it was to drive the car, and a secondary driver, whose job it was to monitor the actions of the first driver and intervene when the first driver made an error. The car architectures of central interest were human primary-machine secondary (``human-machine'') and machine primary-human secondary (``machine-human''). We also included human-human and  machine-machine architectures for comparison. Study 2 compared the human-machine and machine-human shared control cars with two different baseline cars: a standard car, which is exclusively driven by a human, and a fully autonomous car, which is exclusively driven by a machine. In Study 3, we used the same car regimes as in Study 2, but the cases were dilemma scenarios in which the drivers had to choose between crashing into a single pedestrian and crashing into five pedestrians. Across all studies and cases, there were no significant effects of question (blame or causal responsibility), so the data was collapsed across those variables for the analysis.

In all studies, in Bad Intervention cases, two predictors were entered into a regression with rating as the outcome variable: (1) whether or not the driver made an error and (2) driver type (human or machine). The main finding is that across all three studies, whether or not the driver made an error was a significant predictor of ratings, whereas driver type was not (for details, see Table \ref{tab:regression}). In other words, a driver that unnecessarily  intervened, leading to the death of a pedestrian was blamed more than  a driver that operated on the correct course -- regardless of whether the driver was a human or machine.

In Missed Intervention cases, blame and responsibility judgments cannot depend on whether or not a driver made an error because both drivers make errors in these cases. The main finding from these cases is that driver type -- whether the driver is a human or machine -- has a significant impact on ratings in Studies 1 and 2. (There is no effect of driver type on ratings in Study 3, the dilemma cases. We come back to this point in the Discussion.) Specifically, in these shared-control scenarios, where both human and machine have made errors, the machine driver is consistently blamed \textit{less} than the human driver (see Figure \ref{fig:mainresults}).

This human-machine difference appears to be driven by a reduction in the blame attributed to machines when there is a human in the loop. This is evident when comparing both the human-machine and machine-human instances of shared control to the machine-machine scenario. Note that the behaviors in these scenarios are identical, but how much a machine is blamed depends on whether it is sharing control with a human or operating both the primary and secondary driver role.  When the machine is the primary driver, it is held less blameworthy when its secondary driver is a human ($M=57.22$) compared to when the secondary driver is also the machine ($M=68.02$), $t(760.6)=5.05$, $p<.0001$. Similarly, when the machine is the secondary driver, it is held less blameworthy when its primary driver is a human ($M=53.45$), compared to when the primary driver is also the machine, $t(722.77) = -6.6042$, $p<.0001$. The drop in machine responsibility in cases where a human and machine share control of the car is also verified in Study 2, by comparing blame to the machine in the shared control cases with blame to the machine in the Fully Autonomous car. In each case, blame to the machine in the shared control case is significantly lower than blame to the machine in the Fully Autonomous car: Fully Autonomous ($M= 75.56$) vs. Machine-Human ($M= 59.50$), $t(754.63) = -7.3885$, $p<.0001$; vs. Human-Machine ($M= 48.51$), $t(745.06) = 11.676$, $p<.0001$. 

\begin{figure}[!ht]
\centering
\includegraphics[width=1.0\linewidth]{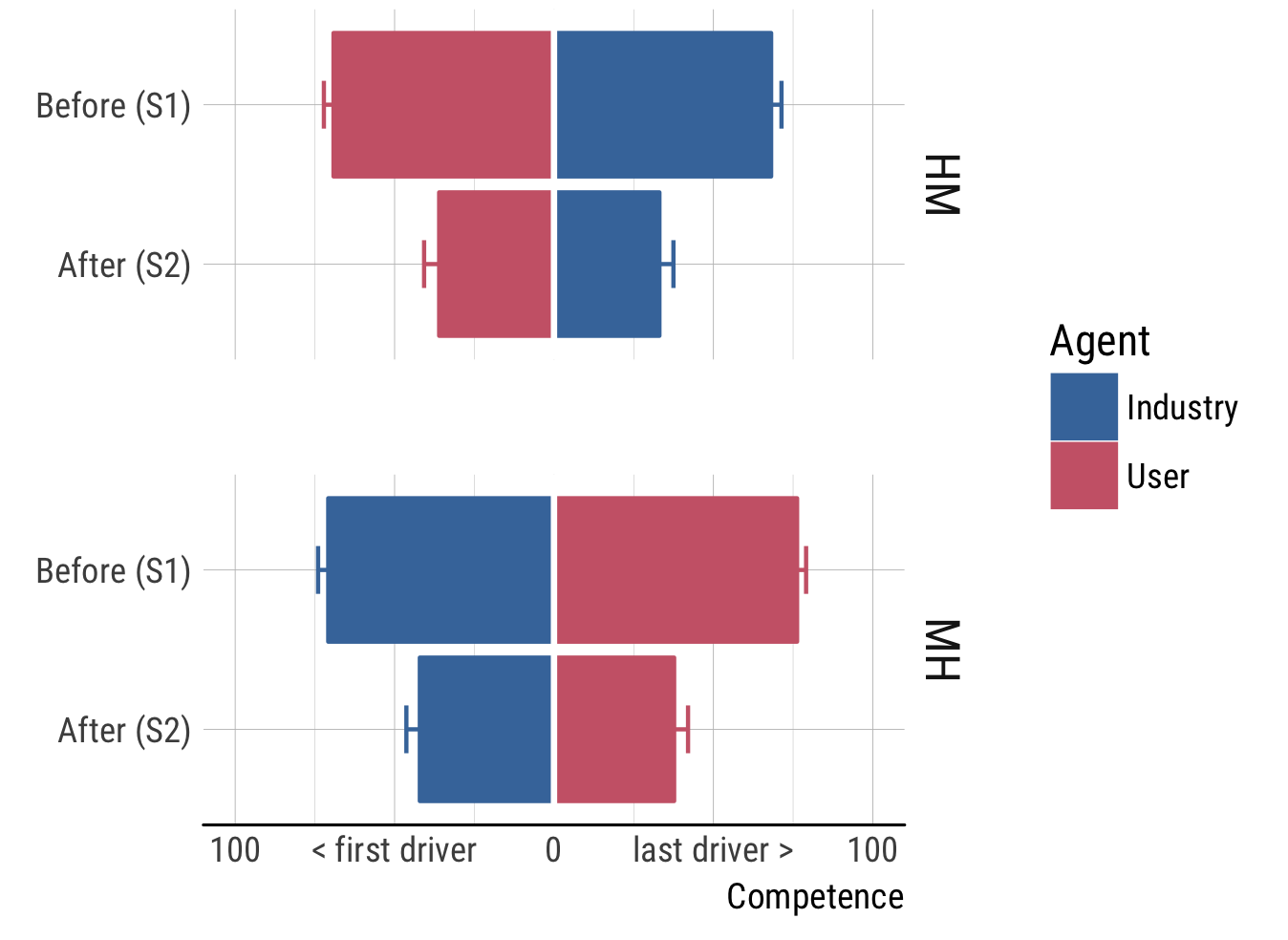}
\caption{\textbf{Competence ratings from studies 1 and 2, in the Missed Intervention cases, for the two critical regimes (human-machine and machine-human).} In Study 1 (S1), ratings were collected before participants read about the accident. In Study 2 (S2), they were collected after. For Industry, ratings of car and company are aggregated (collectively referred to as Industry, henceforth). The x-axis labeling of first driver refers to the main driver, and the last driver refers to the secondary driver in dual-agent cars. Industry and User ratings are shown in blue, and red, respectively. User and Industry receive similar competence ratings in each case. Ratings of both User and Industry drop at the same rate when the question is asked after the scenario is presented.}
\label{fig:mediation}
\end{figure}

Importantly, the effect observed in the Missed Intervention cases  did not reflect a general tendency to blame the machine less: When only one agent makes an error (the Bad Intervention cases), an erring machine is blamed at the same level as an erring human. That is, there is no effect of agent type on blame judgments for cases where humans and machines share control in the Bad Intervention cases (Human: $M=53.36$; Machine: $M=49.78$; $t(1586) = -1.6391$, $p = 0.1014$). Finally, assessments of competence did not mediate blame judgments after a missed intervention (see Figure \ref{fig:mediation}) -- humans and machines were rated as equally competent, both when these ratings were collected before reading about the accident (Study 1) or after (Study 2).

Interestingly, for both types of mistakes (Bad interventions or Missed Interventions), some amount of blame was attributed to machines and humans who were not even in the decision-making loop (in the Regular Car and Fully Autonomous cases, respectively). 
This suggests that passengers in fully autonomous vehicles may be perceived as implicitly accepting some degree of responsibility by even getting in the car at all. Likewise, traditional (non-AI) cars and the car companies that sell them also are perceived as being somewhat liable for the bad outcomes of crashes.

\section{Discussion}

Our central finding was that in cases where a human and a machine share control of the car, less blame is attributed to the machine when both drivers make errors. The first deadly Tesla crash (which killed the car’s occupant, mentioned in the Introduction) was similar in structure to our Missed Intervention cases.  In that case both the machine primary driver and the human secondary driver should have taken action (braking to avoid a truck making a left turn across traffic) and neither driver did.  Our results suggests that the public response that occurred to the crash -- one that focused attention on an unfounded rumor that the driver was exceedingly negligent -- is likely to generalize to other dual-error Missed Intervention-style cases, shifting blame away from the machine and towards the human.  Moreover, the convergence of our results with this real-world public reaction seems to suggest that while we employed stylized, simplified vignettes in our research, our findings show external validity.  

Our central finding (diminished blame to the machine in dual-error cases) leads us to believe that, while there may be many psychological barriers to self-driving car adoption \citep{shariff2017psychological}, public over-reaction to dual-error cases is not likely to be one of them.  In fact, we should perhaps be concerned about public \textit{under}-reaction. Because the public are less likely to see the machine as being at fault in dual-error cases like the Tesla crash, the sort of public pressure that drives regulation might be lacking.  For instance, if we were to allow the regulation of semi-autonomous vehicles to take place through jury-based court-room decisions, we expect that juries will be biased to absolve the car manufacturer of blame in dual-error cases, thereby failing to put sufficient pressure on manufacturers to improve car designs. In fact, we have been in a similar situation before. Prior to the 1960s, car manufacturers enjoyed a large amount of liberty from liability when a car's occupant was harmed in a crash (because blame in car crashes was attributed to the driver's error or negligence). Top-down regulation was necessary to introduce the concept of ``crash worthiness'' into the legal system, that is, that cars should be designed in such a way to minimize injury to occupants when a crash occurs. Only following these laws were car manufacturers forced to improve their designs \citep{nader1965unsafe}. Here, too, top-down regulation of semi-autonomous car safety might be needed to correct a public under-reaction to crashes in shared-control cases.  What, exactly, the safety standard should be is still an open question, however.

If our data identifies a source of possible public over-reaction, it is for cars with a human primary driver and a machine secondary driver in Bad Intervention-style cases. These are the only cases we identified where the car receives more blame than the human. After all, in these kinds of cars, it is possible for the machine to make an error that it is not the human's responsibility to correct.  That is, the architecture of this car is as follows: the machine's job, as the secondary driver, is to correct any mistakes the human may make, but if the machine makes a mistake, it is not the human's job to correct it. (This is not true for cars with the machine primary driver and a human secondary driver. In this kind of car, when the machine makes an error, it is always the human's job to correct it.) Therefore, the car is blamed more than the human in Bad Intervention cases. It seems possible that these sorts of cars may generate widespread public concern once we see instances of Bad Intervention-style crashes in human-machine car regimes. This could potentially slow the transition to fully autonomous vehicles if this reaction is not anticipated and managed appropriately in public discourse and legal regulation.  Moreover, manufacturers that are working to release cars with a machine secondary driver should plan appropriately for the likely legal fall-out for these unique cases where a machine driver receives more blame than a human.

Our data portends the sort of reaction we can expect to semi-autonomous car crashes at the societal level (for example, through public reaction and pressure to regulate). Once we begin to see societal-level responses to semi-autonomous cars, that reaction may shape incentives for individual actors. For example, people may want to opt into systems that are designed such that, in the event of a crash, the majority public response will be to blame the machine. Worse yet, people may train themselves to drive in a way that, if they crash, the blame is likely to fall to the machine (for instance, by not attempting to correct a mistake that is made by a machine over-ride). This sort of incentive shaping may already be happening in the legal domain. Judges who make decisions about whether to release a person from custody between arrest and a trial frequently rely on actuarial risk assessment tables to help make their decision. Some suspect that judges may overly rely on the tables as a way of diminishing their responsibility if a released person commits a crime. Recent attention generated in response to such a case focused on the role of the algorithm rather than the judge \citep{Westervelt2017bail}, indicating that the possibility of incentive shaping in the legal domain is not so far-fetched.

Studies 1 and 2 looked at blame and causal responsibility attribution in cases where one or both drivers made errors. Study 3 looked at dilemma scenarios where the drivers faced the choice of running over either one or five pedestrians.  While there is, in some sense, an ``optimal'' outcome in these cases (corresponding to saving more lives), it is not obvious that it would (for example) count as an error to refuse to swerve away from five pedestrians into a pedestrian that was previously unthreatened. In fact, the German Ethics Commission on Automated and Connected Driving recently released a report indicating that programming cars to trade off lives in this way would be prohibited. The report states: ``It is also prohibited to offset victims against one another. [$\ldots$] Those parties involved in the generation of mobility risks must not sacrifice non-involved parties.'' Even though participants in previous studies prefer to sacrifice one person who was previously not involved than five (e.g., \citep{bonnefon2016autonomous}), the German Ethics Commission's decision underscores the fact that trading off lives in dilemma situations can be particularly fraught. For this reason, and for continuity with previous work on the ethics of self-driving cars \citep{bonnefon2016autonomous,li2016trolley} and in moral psychology more generally \citep{mikhail2011elements,greene2014moral}, we chose to investigate dilemma situations. Our findings that there is no effect of driver type in these cases underscores the fact that findings about how blame and responsibility are attributed after a crash may not generalize to less-clear dilemma scenarios.

Some of our results fall in line with previous work on the psychology of causal inference.  In Bad Intervention cases, the primary driver (be it human or machine) makes a correct decision to keep the car on course, which will avoid a pedestrian. Following this, the secondary driver makes the decision to swerve the car into the pedestrian. Our data show that the secondary driver (the one that makes a mistake) is considered more causally responsible than the first driver. It is well established that judgments of causal responsibility are impacted by violations of statistical and moral norms \citep{alicke2000culpable,gerstenberg2015whether,hitchcock2009cause,kominsky2015causal}, and a mistake seems to count as such a violation. That is, if something unusual or counter-normative happens, that event is more likely to be seen as a cause of some effect than another event that is typical or norm-conforming. 

Other results, though, are more surprising in light of previous work on the psychological of causal inference.  In each of the three studies, subjects gave substantial blame and responsibility to machine drivers, often at levels not different from the levels of blame they ascribed to human drivers in similar roles. Research on the psychology of causal attribution suggests that voluntary causes (causes created by agents) are better causal explanations than physical causes \citep{hart1985causation}, which would imply that more causal responsibility would be attributed to the human over the machine driver. On the other hand, it remains an open question whether an AI that is operating a car counts as a physical cause, an agent, something in between, or something else entirely \citep{gray2007dimensions,weisman2017rethinking}. Future work should investigate the mental properties attributed to an AI that controls a car both in conjunction with a human or alone. Understanding the sort of mind we perceive as dwelling inside an AI may help us understand and predict how blame and causal responsibility will be attributed to it \citep{gray2012mind}.

There is another reason why it is particularly surprising that the machine drivers are attributed similar levels of blame to human drivers in the same role.  There are various ways that humans express moral condemnation. For example, we may call an action morally wrong, say that a moral agent has a bad character, or judge that an agent is blameworthy. Judgments of blame typically track judgments of willingness to punish the perpetrator \citep{cushman2008crime,cushman2015deconstructing}. Are the participants in our study expressing that some punishment is due to the machine driver of the car, whatever that may mean? Alternately, is it possible that subjects' expressions of blame indicate that some entity is deserving of punishment that represents the machine (the company, or a human representative of the company, such as the CEO). The similar blame judgments given to the car and the car's representatives (company, programmer) perhaps support this possibility. Finally, it is possible that participants ascribe only non-moral blame to the machine, in the sense of being responsible but not in a moral sense. We may say that a forest fire is to blame for displacing residents from their homes without implying that punishment is due to anyone at all.

Following these studies, the reason that subjects blame machine drivers less than human drivers in Missed Intervention cases remains an open question. One possibility is that this pattern is a feature of our uncertainty how to perceive the agential status of machines and what it means to blame a machine at all.  Once machines are a more common element in our moral world and we interact with machines as moral actors, will this effect change? Or will this finding be a lasting hallmark of the cognitive psychology of human-machine interaction?

A final open question concerns whether the effects we report here will generalize to other cases of human-machine interaction. Already we see fruitful human-machine partnerships emerging with judges, doctors, military personnel, factory workers, artists, and financial analyzes, just to name a few. We conjecture that we may see the patterns we report here in domains other than autonomous vehicles, though each domain will have its own complications and quirks as machines begin to become more subtly integrated in our personal and professional lives.

\section*{Methods}
\subsection*{Study 1}

The data was collected in September 2017 from $786$ participants (USA residents) recruited from the Mechanical Turk platform. Participants were uniformly randomly allocated to one of four conditions. Conditions varied the car type (human-human, human-machine, machine-human, and machine-machine) in a 4-level between-subjects design. Each car had a primary driver, whose job it was to drive the car, and a secondary driver, whose job it was to monitor the actions of the first driver and intervene when the first driver made an error. In each condition, participants first read a description of the car, and were then asked to attribute competence to each of the two drivers on an 100-point scale anchored at ``not competent'' and ``very competent''. Participants then read two scenarios (presented in a random order), one Bad Intervention case and one Missed Intervention Case. In each case, the main driver decided to do nothing (kept the car on its track). The secondary driver intervened in the Bad Intervention Cases (steering the car into another lane and killing a pedestrian there). The secondary driver did not intervene (kept the car on its track) in the Missed Intervention cases (and killed a pedestrian walking in that lane, rather than swerving and killing no one). After each scenario, participants were asked to indicate (on an 100-point scale) to what extent they thought each driver was blame-worthy (from ``not blame-worthy'' to ``very blame-worthy'') and to what degree each of these two agents caused the death of the pedestrian (from ``very little'' to ``very much'').  Questions were presented in a randomized order. (See Supplemental Materials for text of the vignettes and questions). At the end of the surveys, participants provided basic demographic information (e.g., age, sex, income, political views).

\subsection*{Study 2}

The data was collected in May 2017 from $779$ participants (USA residents) recruited from the Mechanical Turk platform. Participants were uniformly randomly allocated to one of eight conditions. Conditions varied the car type (human only, human-machine, machine-human, and machine only) and the industry representative (car and company), in a 4x2 between-subject multi-factorial design. In each condition, participants read two scenarios (presented in a random order), one Bad Intervention case and one Missed Intervention case. The two-driver scenarios were identical to those in Study 1. In the single-driver scenarios, the sole driver in the Bad Intervention cases steered the car off its track (killing a pedestrian) rather than keeping the car on track and killing no one. The sole driver in the Missed Intervention cases kept the car on its track (killing a pedestrian) rather than swerving into the adjacent lane and killing no one. After each scenario, participants were asked to attribute causal responsibility, blameworthiness, and competence to two agents: the human in the car and a representative of the car (the car itself or the manufacturing company of the car, depending on the condition). All other features of Study 2 were the same at those in Study 1.

\subsection*{Study 3}

The $973$ participants (USA residents only) were recruited from the Mechanical Turk platform. Participants were uniformly randomly allocated to one of twelve conditions. Conditions varied the car type (human only, human-machine, machine-human, and machine only) and the industry representative (car, company, and programmer), in a 4x3 between-subject multi-factorial design. In each condition, participants read two scenarios (presented in a random order), one Bad Intervention case and one Missed Intervention case. These scenarios were identical to those presented in Studies 1 and 2, except that the drivers in the Bad Intervention cases swerved out of the way of one pedestrian and killed five and the drivers in the Missed Intervention cases kept the car on track to kill five pedestrians rather than swerving to kill one. After each scenario, participants were asked to attribute causal responsibility and blameworthiness to two agents: the human in the car and a representative of the car (the car itself, the company, or the programmer, depending on the condition). All other features of Study 3 were identical to those of Study 2.

In all studies, to get our final sample, we excluded any subject who did not (i) complete all measures within the survey, (ii) transcribe (near-perfectly) a 169-character paragraph from an image (used to exclude non-serious Turkers), and (iii) have unique TurkID per study (all records with a recurring MTurk ID were excluded).

\bibliographystyle{plainnat}
\bibliography{sample}

\clearpage
~\\~\\

{\Huge \bfseries Supplemental Information} \\~\\

\section*{A. Full Results}
Tables \ref{tab:s1}, \ref{tab:s2}, and \ref{tab:s3} show the mean and the $95\%$ confidence interval for each condition for studies 1, 2 and 3. 

\begin{table}[ht]
\centering
\caption{Study 1 results. Mean and $95\%$ confidence interval for each condition.} \label{tab:s1}
\begin{tabular}{|l|l|l|l|l|r|l|}
  \hline
Scenario Type & Car & Agent & Agent & Agent is the & Mean & $95\%$ CI \\ 
 & Type &  &  Erred? & last driver? &  &  \\ 
  \hline
Bad Intervention & HH & User & No & No & 21.35 & 17.98--24.72 \\ 
  Bad Intervention & HH & User & Yes & Yes & 84.17 & 81.18--87.17 \\ 
  Bad Intervention & HM & Industry & Yes & Yes & 83.85 & 81.05--86.64 \\ 
  Bad Intervention & HM & User & No & No & 19.30 & 16.37--22.23 \\ 
  Bad Intervention & MH & Industry & No & No & 16.21 & 13.59--18.82 \\ 
  Bad Intervention & MH & User & Yes & Yes & 86.90 & 84.46--89.34 \\ 
  Bad Intervention & MM & Industry & No & No & 22.10 & 18.86--25.34 \\ 
  Bad Intervention & MM & Industry & Yes & Yes & 81.09 & 78.05--84.12 \\ 
  Missed Intervention & HH & User & Yes & No & 73.53 & 70.21--76.86 \\ 
  Missed Intervention & HH & User & Yes & Yes & 73.96 & 70.95--76.98 \\ 
  Missed Intervention & HM & Industry & Yes & Yes & 53.45 & 49.81--57.09 \\ 
  Missed Intervention & HM & User & Yes & No & 77.65 & 74.63--80.67 \\ 
  Missed Intervention & MH & Industry & Yes & No & 57.22 & 53.73--60.7 \\ 
  Missed Intervention & MH & User & Yes & Yes & 72.73 & 69.72--75.75 \\ 
  Missed Intervention & MM & Industry & Yes & No & 67.51 & 64.19--70.82 \\ 
  Missed Intervention & MM & Industry & Yes & Yes & 68.53 & 65.23--71.82 \\ 
   \hline
\end{tabular}
\end{table}

\begin{table}[ht]
\centering
\caption{Study 2 results. Mean and $95\%$ confidence interval for each condition.} \label{tab:s2}
\begin{tabular}{|l|l|l|l|l|r|l|}
  \hline
Scenario Type & Car & Agent & Agent & Agent is the & Mean & $95\%$ CI \\ 
 & Type &  &  Erred? & last driver? &  &  \\ 
  \hline
Bad Intervention & H & Industry & No & No & 17.56 & 14.63--20.49 \\ 
  Bad Intervention & H & User & Yes & Yes & 84.54 & 81.68--87.4 \\ 
  Bad Intervention & HM & Industry & Yes & Yes & 78.15 & 74.88--81.42 \\ 
  Bad Intervention & HM & User & No & No & 23.95 & 20.76--27.14 \\ 
  Bad Intervention & M & Industry & Yes & Yes & 80.61 & 77.94--83.28 \\ 
  Bad Intervention & M & User & No & No & 29.58 & 26.45--32.7 \\ 
  Bad Intervention & MH & Industry & No & No & 22.14 & 19.15--25.13 \\ 
  Bad Intervention & MH & User & Yes & Yes & 83.06 & 80.21--85.9 \\ 
  Missed Intervention & H & Industry & No & No & 36.30 & 32.9--39.69 \\ 
  Missed Intervention & H & User & Yes & Yes & 75.73 & 72.84--78.62 \\ 
  Missed Intervention & HM & Industry & Yes & Yes & 48.52 & 45--52.03 \\ 
  Missed Intervention & HM & User & Yes & No & 72.77 & 69.53--76.01 \\ 
  Missed Intervention & M & Industry & Yes & Yes & 75.56 & 72.69--78.43 \\ 
  Missed Intervention & M & User & No & No & 32.63 & 29.45--35.82 \\ 
  Missed Intervention & MH & Industry & Yes & No & 59.50 & 56.35--62.65 \\ 
  Missed Intervention & MH & User & Yes & Yes & 70.23 & 67.2--73.27 \\ 
   \hline
\end{tabular}
\end{table}

\begin{table}[ht]
\centering
\caption{Study 3 results. Mean and $95\%$ confidence interval for each condition.} \label{tab:s3}
\begin{tabular}{|l|l|l|l|l|r|l|}
  \hline
Scenario Type & Car & Agent & Agent & Agent is the & Mean & $95\%$ CI \\ 
 & Type &  &  Erred? & last driver? &  &  \\ 
  \hline
Bad Intervention & H & Industry & No & No & 30.59 & 28.58--32.6 \\ 
  Bad Intervention & H & User & Yes & Yes & 35.86 & 33.88--37.85 \\ 
  Bad Intervention & HM & Industry & Yes & Yes & 39.73 & 37.93--41.54 \\ 
  Bad Intervention & HM & User & No & No & 26.44 & 24.92--27.96 \\ 
  Bad Intervention & M & Industry & Yes & Yes & 32.43 & 30.53--34.32 \\ 
  Bad Intervention & M & User & No & No & 18.69 & 17.16--20.21 \\ 
  Bad Intervention & MH & Industry & No & No & 28.09 & 26.48--29.69 \\ 
  Bad Intervention & MH & User & Yes & Yes & 37.99 & 36.21--39.77 \\ 
  Missed Intervention & H & Industry & No & No & 30.72 & 28.78--32.67 \\ 
  Missed Intervention & H & User & Yes & Yes & 38.12 & 36.06--40.18 \\ 
  Missed Intervention & HM & Industry & Yes & Yes & 36.50 & 34.78--38.21 \\ 
  Missed Intervention & HM & User & Yes & No & 39.08 & 37.27--40.89 \\ 
  Missed Intervention & M & Industry & Yes & Yes & 35.50 & 33.56--37.43 \\ 
  Missed Intervention & M & User & No & No & 19.08 & 17.59--20.58 \\ 
  Missed Intervention & MH & Industry & Yes & No & 36.74 & 35.03--38.45 \\ 
  Missed Intervention & MH & User & Yes & Yes & 41.62 & 39.78--43.45 \\ 
   \hline
\end{tabular}
\end{table}
\clearpage
\section*{B. Vignettes}

\subsection*{Study 1}
\begin{figure}[h!]
\centering
\frame{\includegraphics[width=1.0\linewidth]{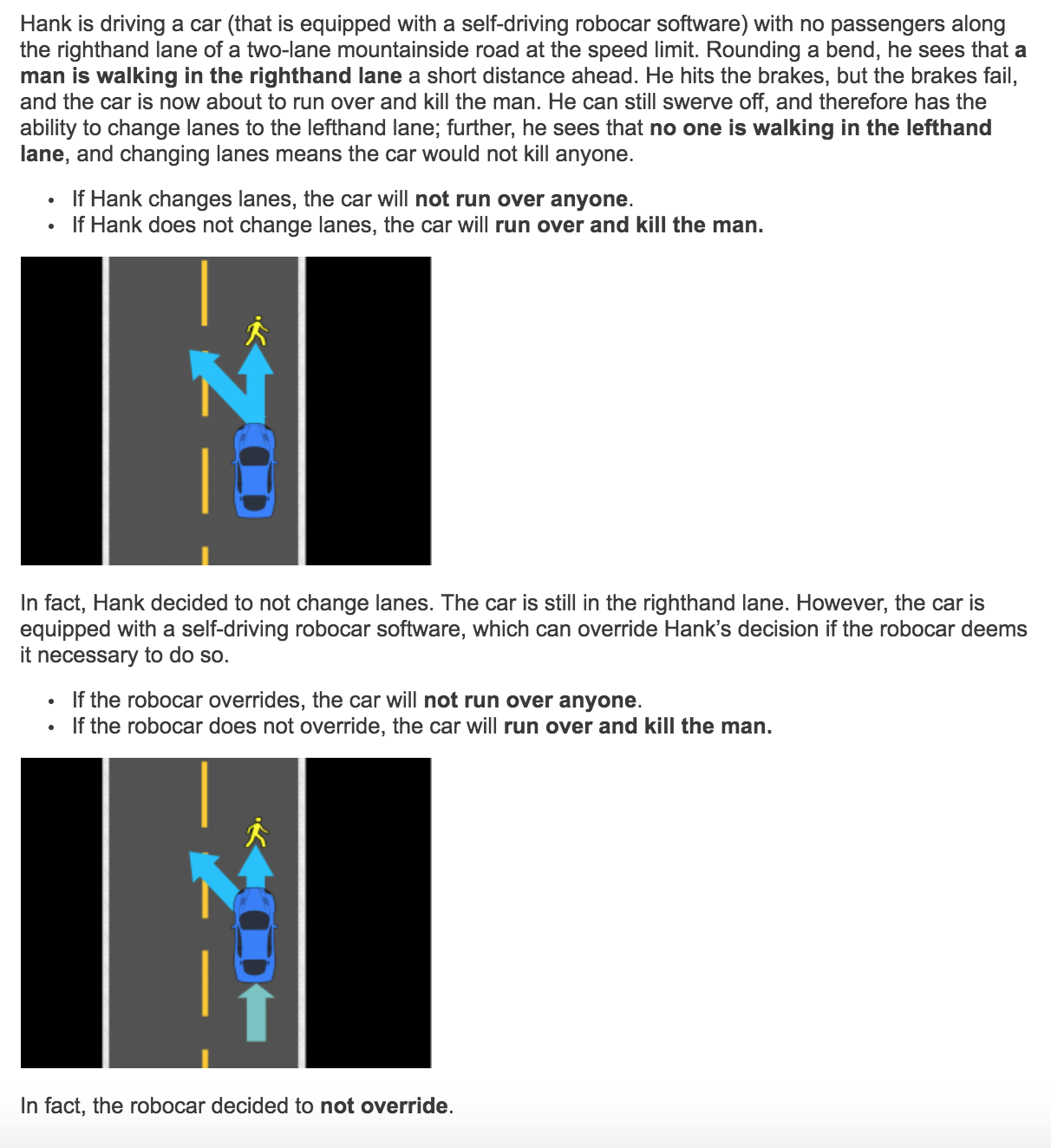}}
\caption{Vignette for Human-Machine with Missed Intervention.}
\label{fig:s1.hm.m}
\end{figure}

\begin{figure}[h!]
\centering
\frame{\includegraphics[width=1.0\linewidth]{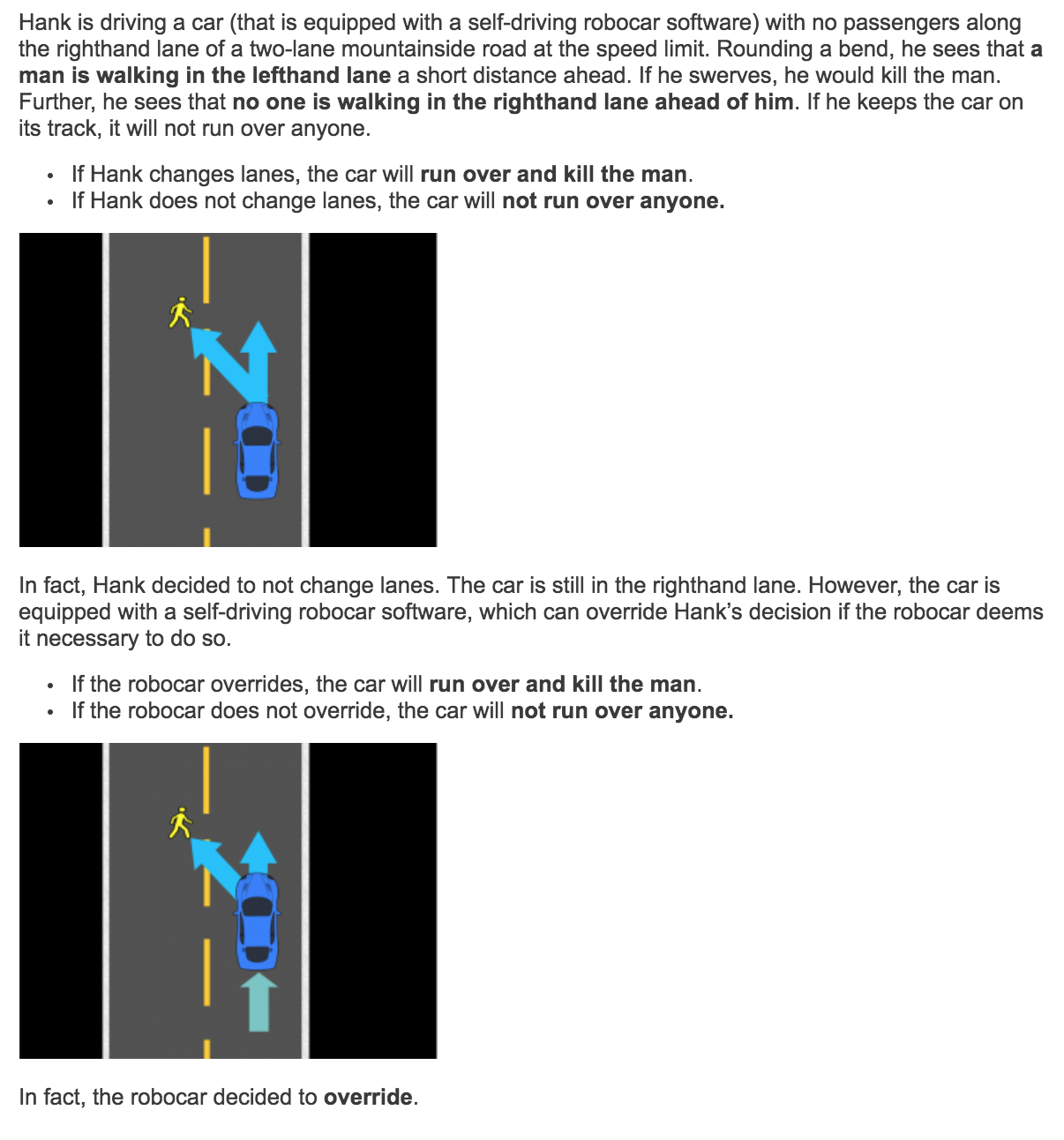}}
\caption{Vignette for Human-Machine with Bad Intervention.}
\label{fig:s1.hm.b}
\end{figure}

\begin{figure}[h!]
\centering
\frame{\includegraphics[width=1.0\linewidth]{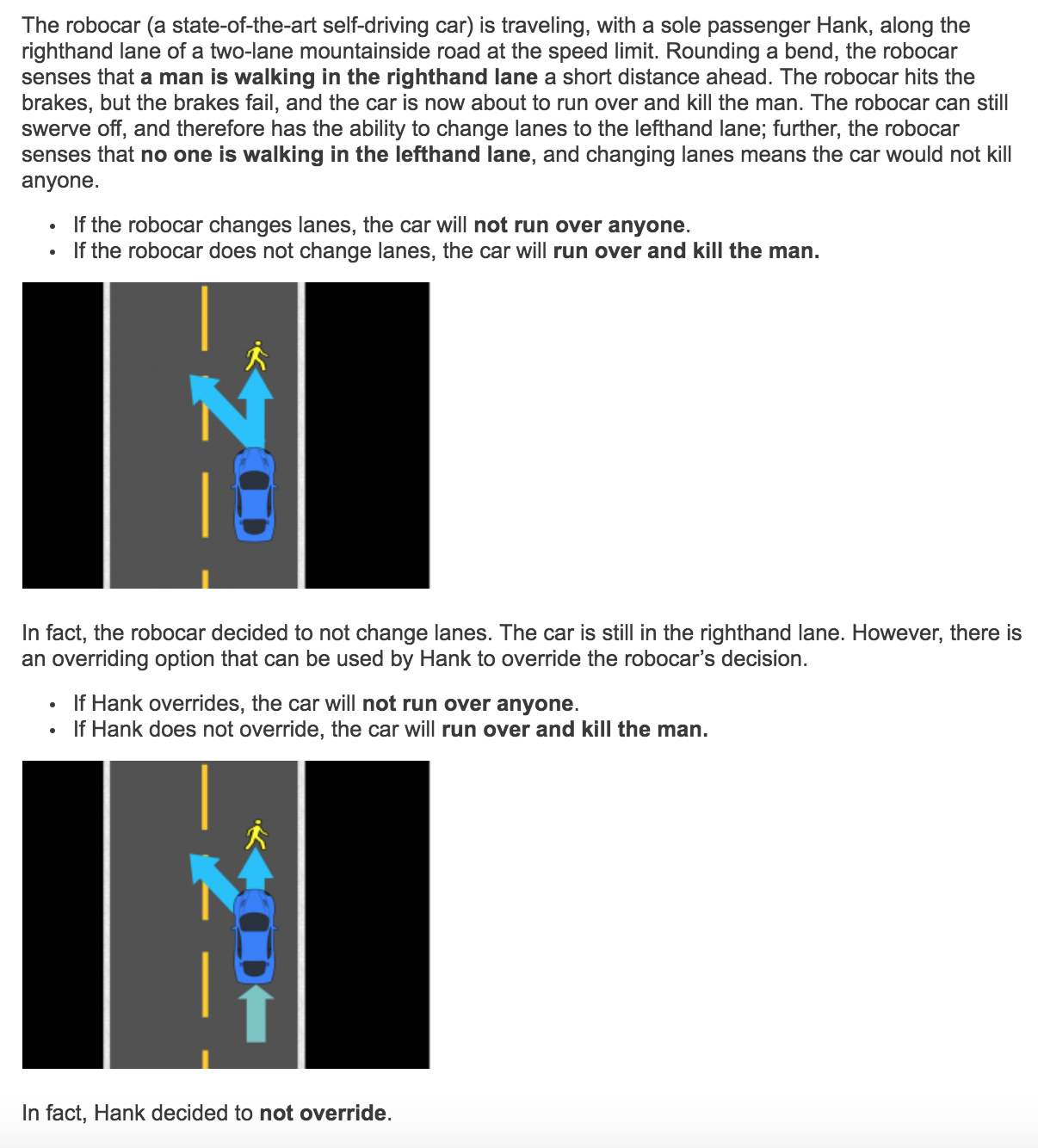}}
\caption{Vignette for Machine-Human with Missed Intervention.}
\label{fig:s1.mh.m}
\end{figure}

\begin{figure}[h!]
\centering
\frame{\includegraphics[width=1.0\linewidth]{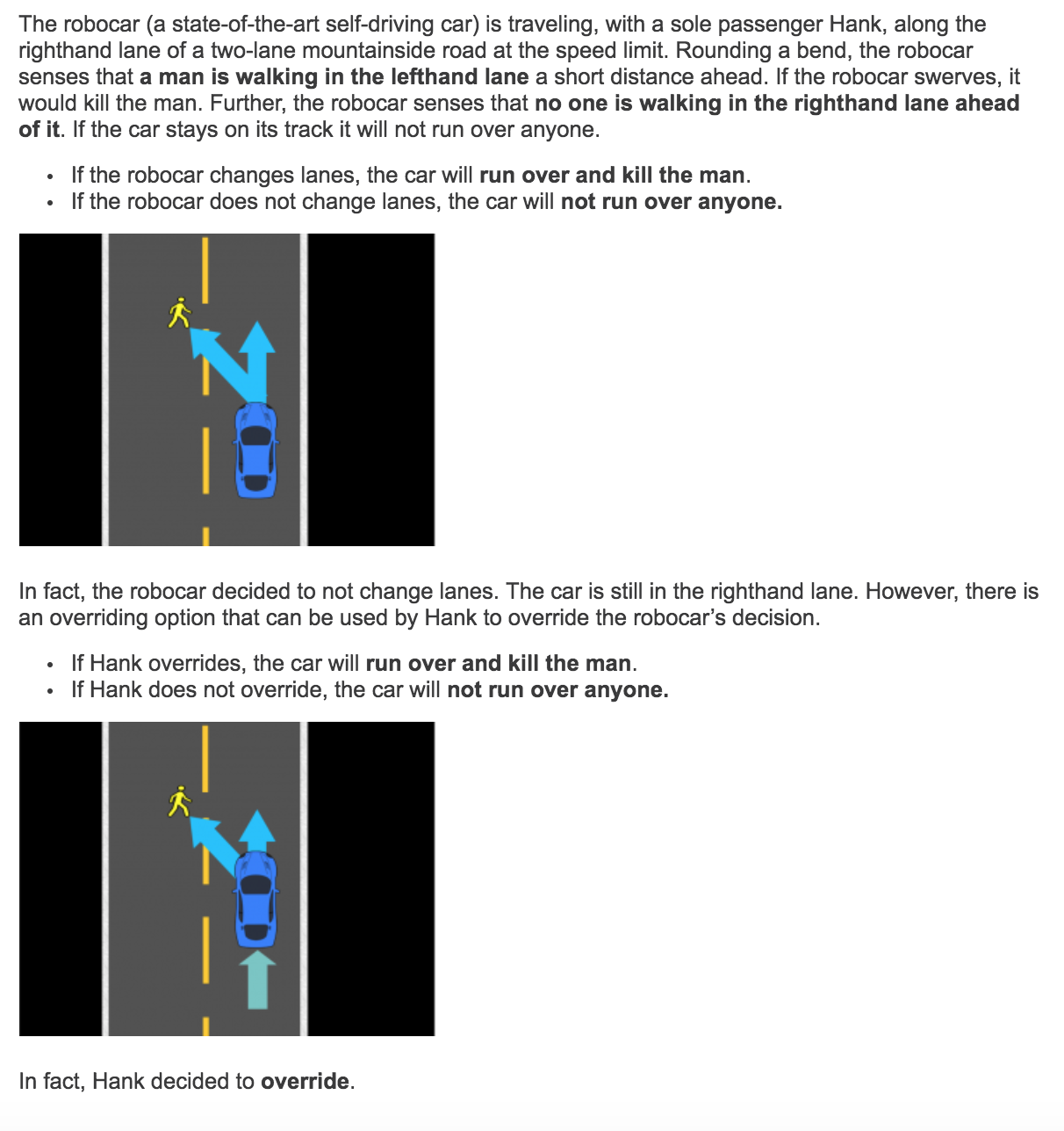}}
\caption{Vignette for Machine-Human with Bad Intervention.}
\label{fig:s1.mh.b}
\end{figure}

\begin{figure}[h!]
\centering
\frame{\includegraphics[width=1.0\linewidth]{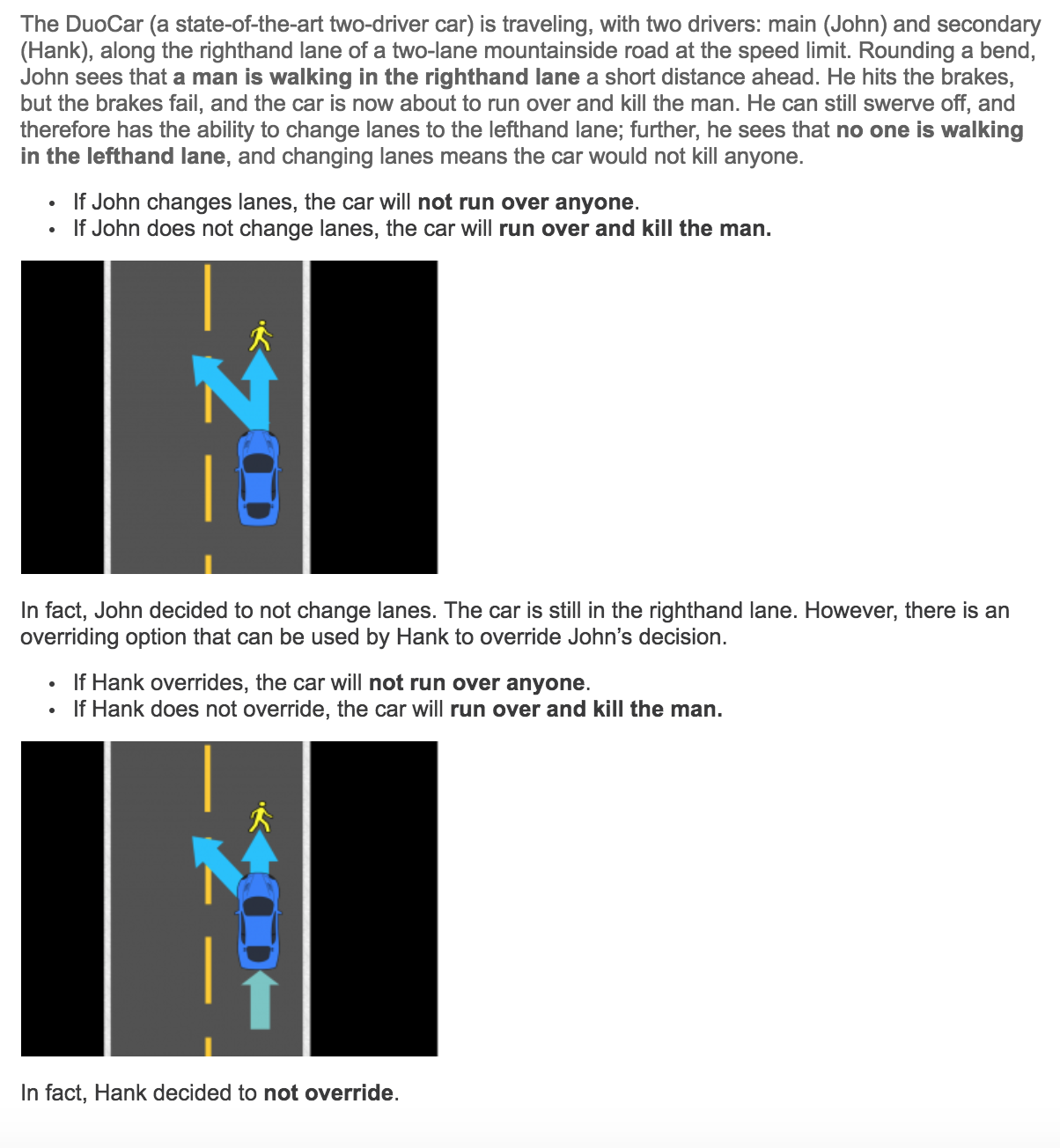}}
\caption{Vignette for Human-Human with Missed Intervention.}
\label{fig:s1.hh.m}
\end{figure}

\begin{figure}[h!]
\centering
\frame{\includegraphics[width=1.0\linewidth]{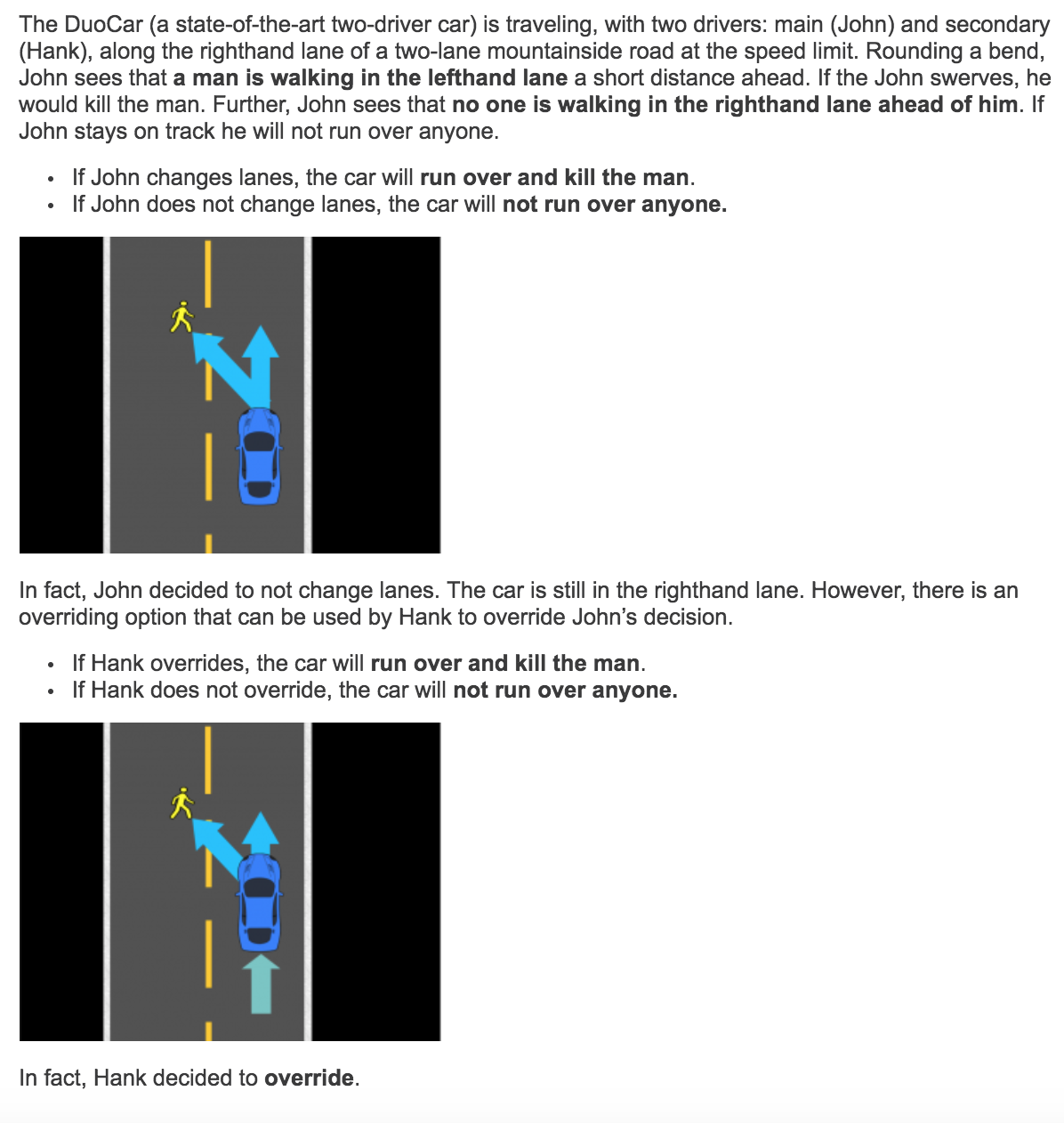}}
\caption{Vignette for Human-Human with Bad Intervention.}
\label{fig:s1.hh.b}
\end{figure}

\begin{figure}[h!]
\centering
\frame{\includegraphics[width=1.0\linewidth]{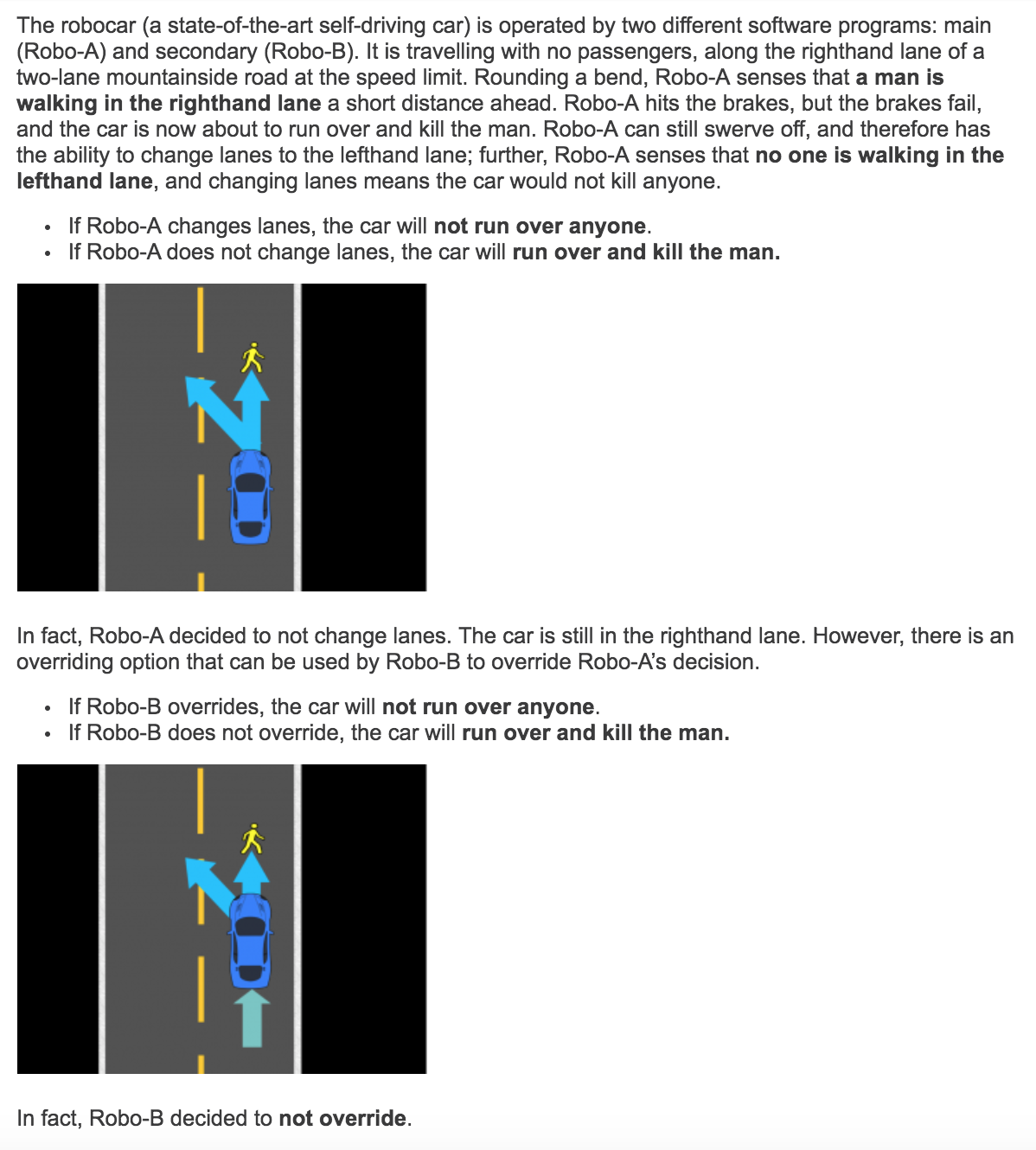}}
\caption{Vignette for Machine-Machine with Missed Intervention.}
\label{fig:s1.mm.m}
\end{figure}

\begin{figure}[h!]
\centering
\frame{\includegraphics[width=1.0\linewidth]{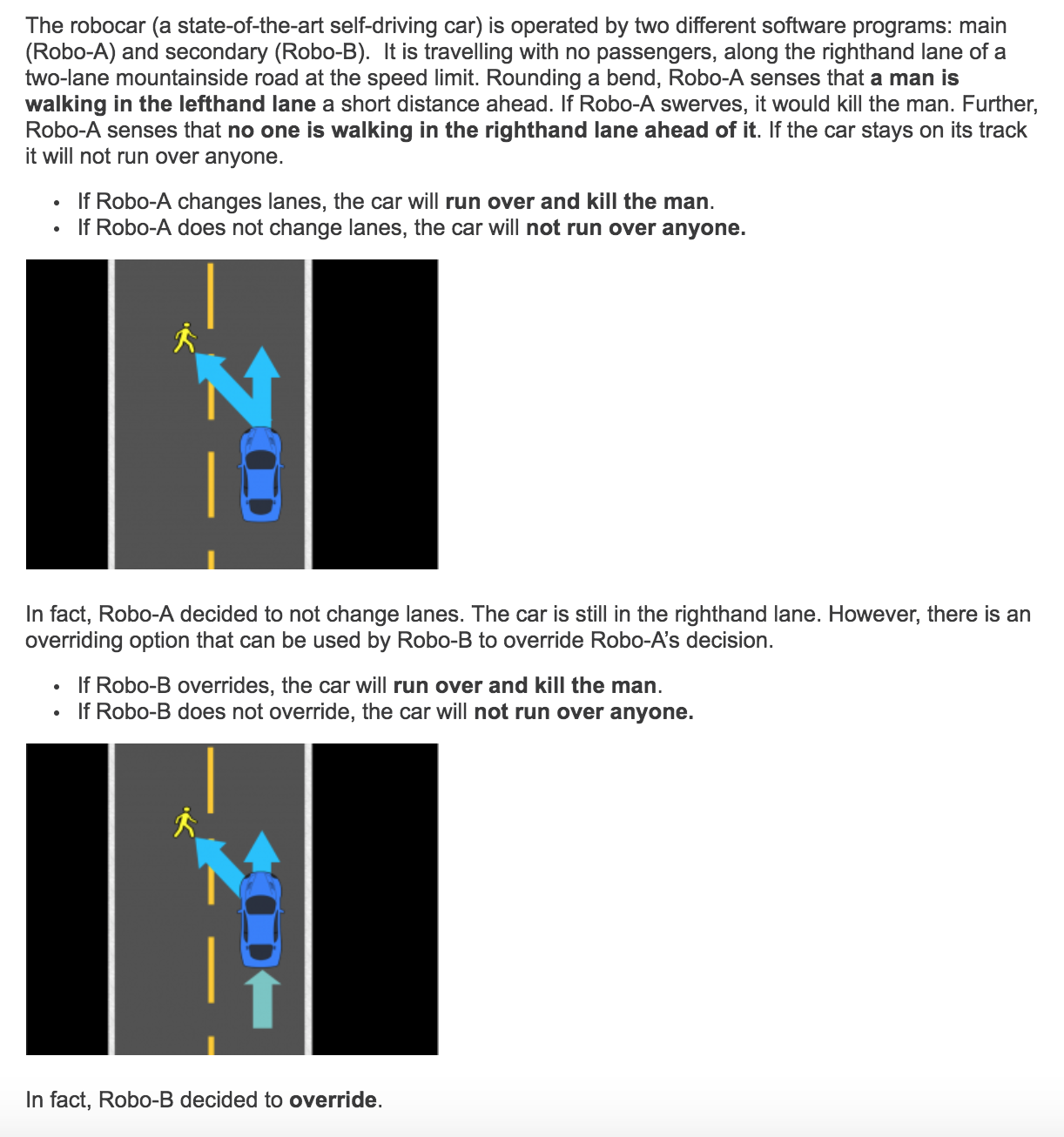}}
\caption{Vignette for Machine-Machine with Bad Intervention.}
\label{fig:s1.mm.b}
\end{figure}

\clearpage

\subsection*{Study 2}
\begin{figure}[h!]
\centering
\frame{\includegraphics[width=1.0\linewidth]{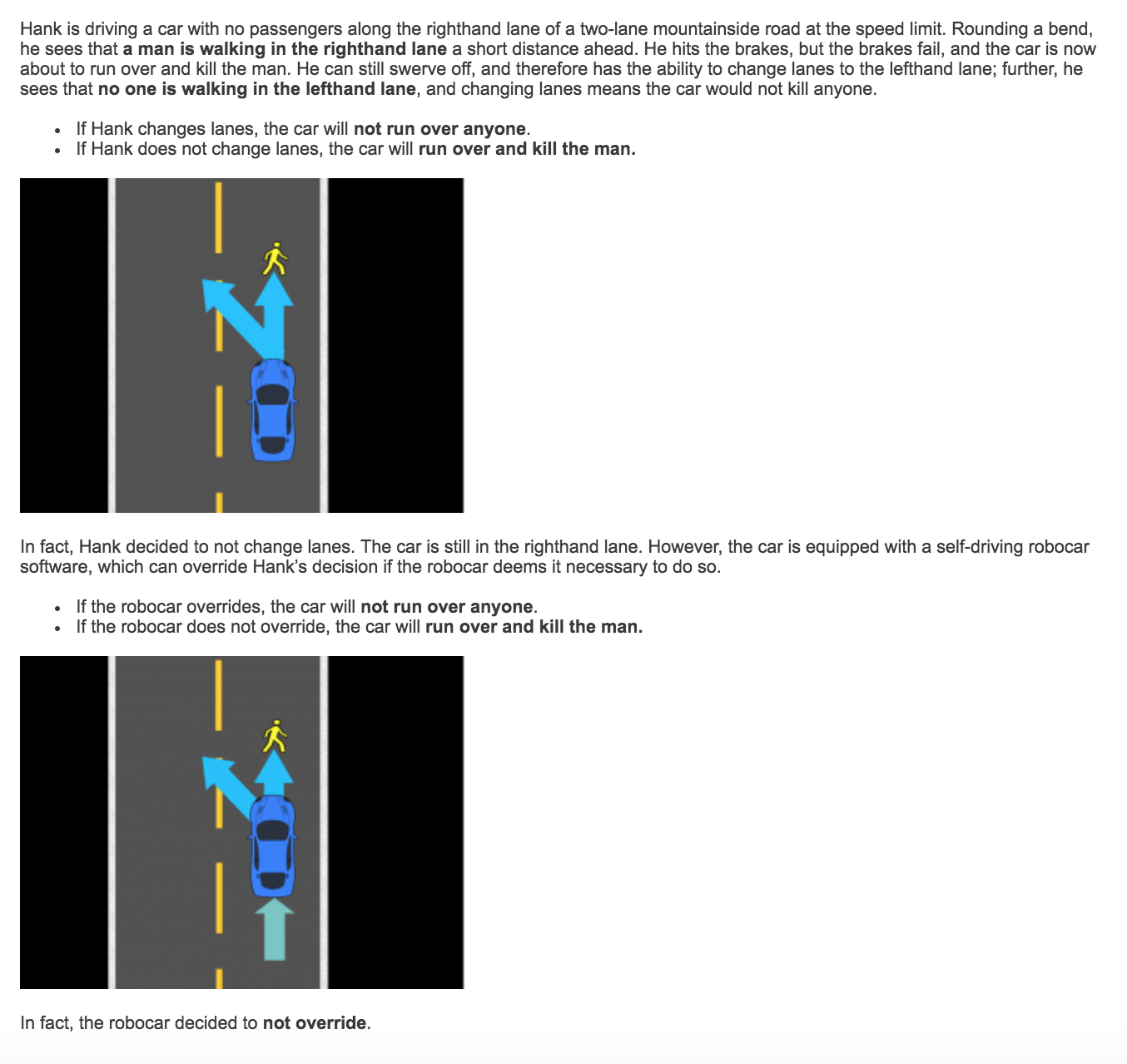}}
\caption{Vignette for Human-Machine with Missed Intervention.}
\label{fig:s2.hm.m}
\end{figure}

\begin{figure}[h!]
\centering
\frame{\includegraphics[width=1.0\linewidth]{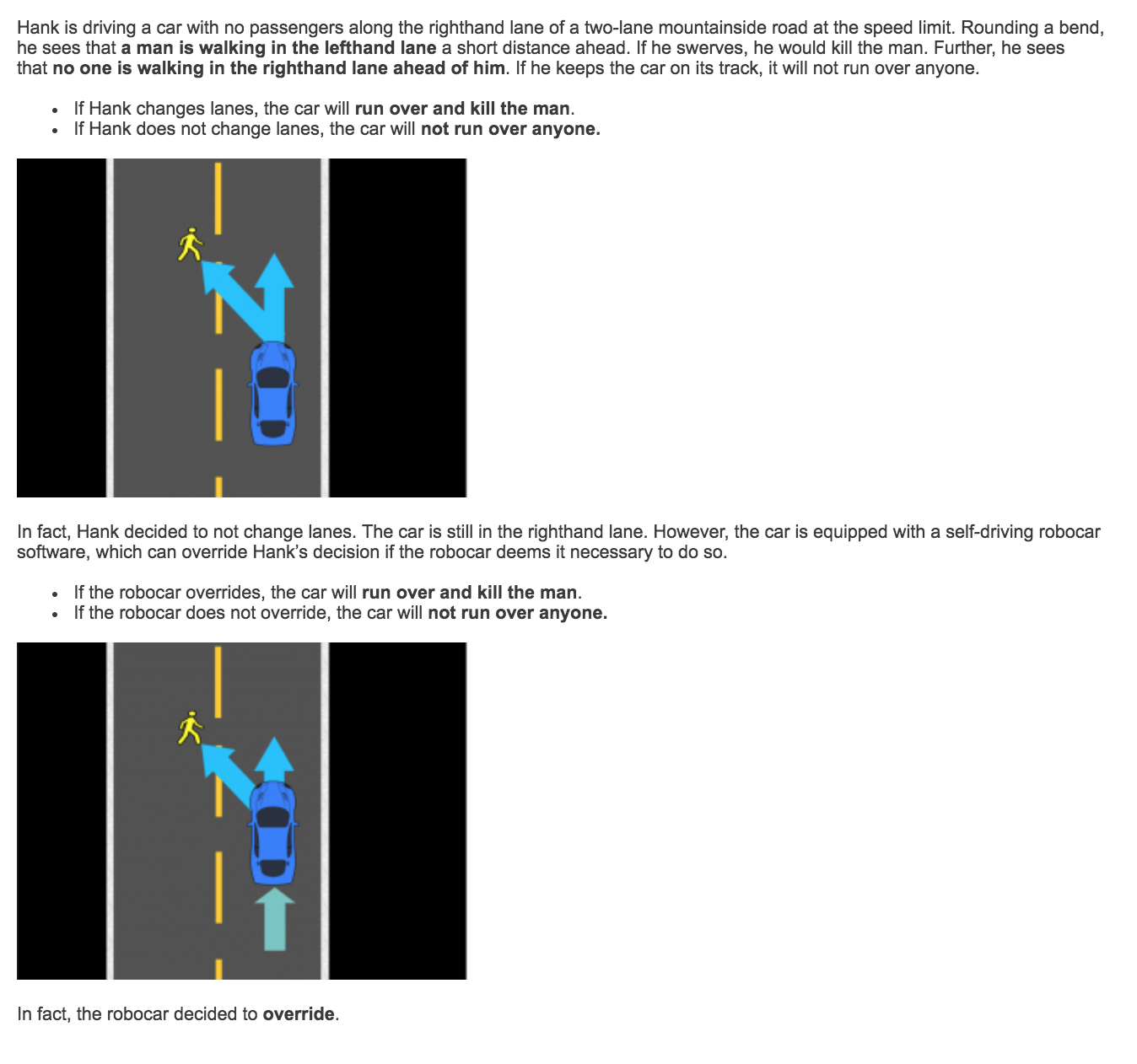}}
\caption{Vignette for Human-Machine with Bad Intervention.}
\label{fig:s2.hm.b}
\end{figure}

\begin{figure}[h!]
\centering
\frame{\includegraphics[width=1.0\linewidth]{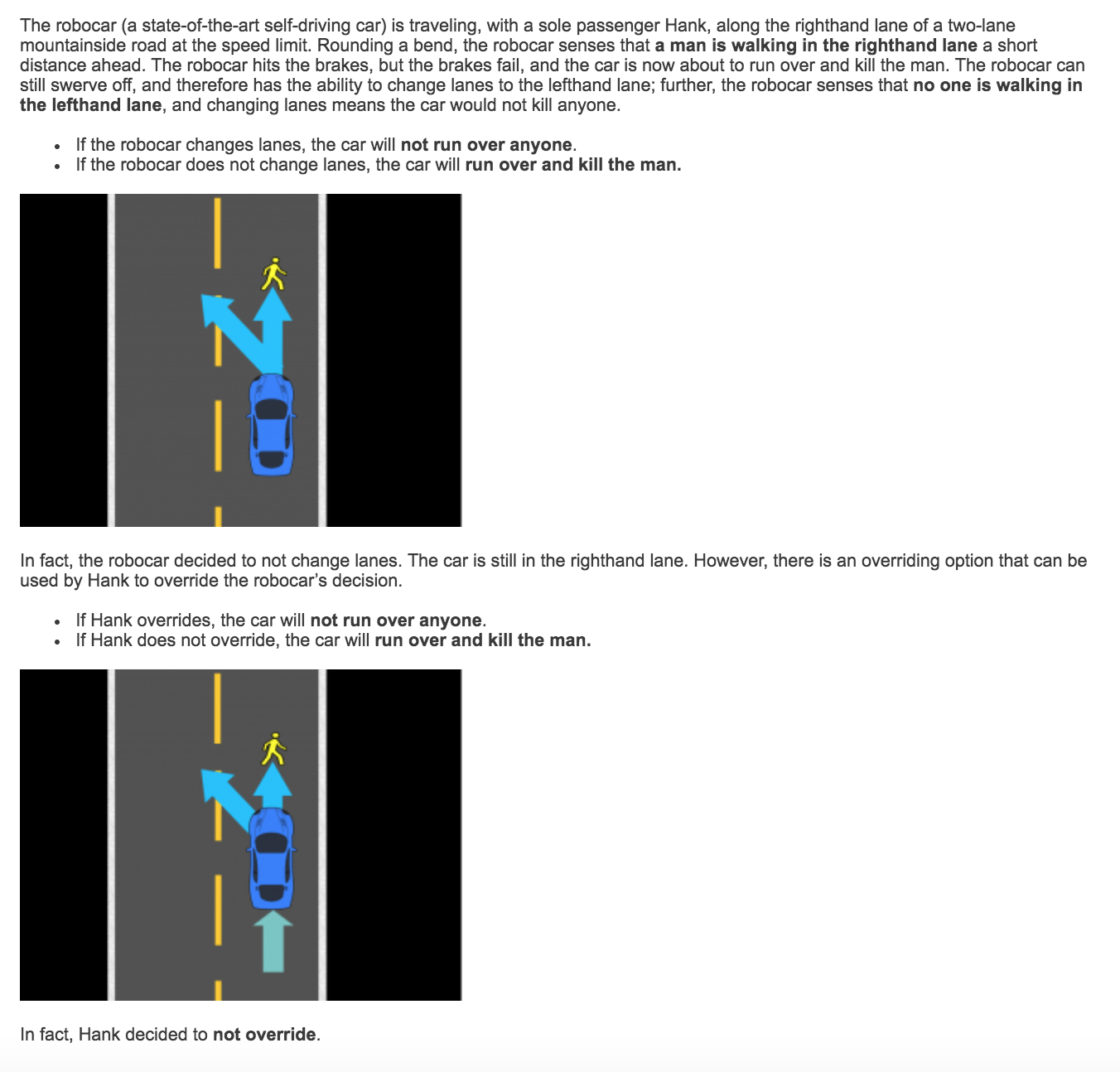}}
\caption{Vignette for Machine-Human with Missed Intervention.}
\label{fig:s2.mh.m}
\end{figure}

\begin{figure}[h!]
\centering
\frame{\includegraphics[width=1.0\linewidth]{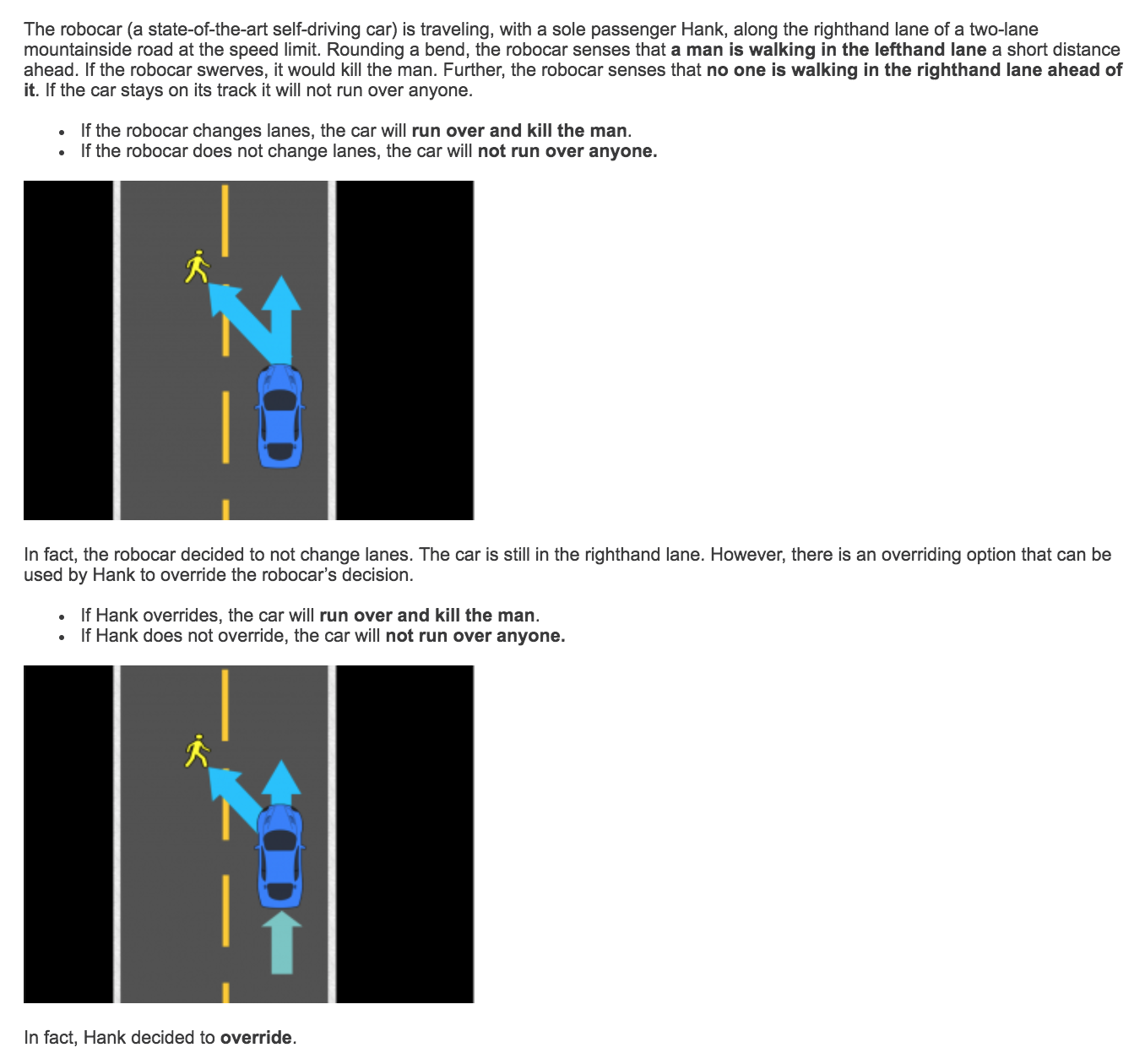}}
\caption{Vignette for Machine-Human with Bad Intervention.}
\label{fig:s2.mh.b}
\end{figure}

\begin{figure}[h!]
\centering
\frame{\includegraphics[width=1.0\linewidth]{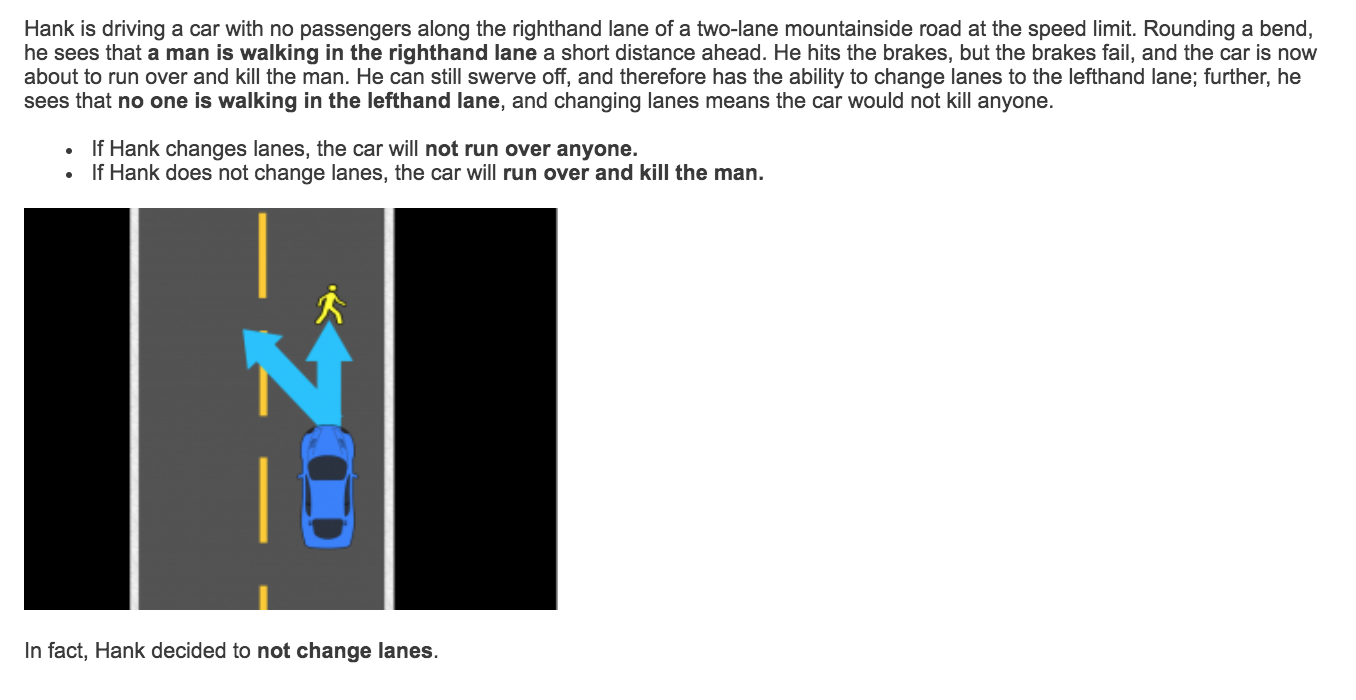}}
\caption{Vignette for Human only (Regular Car) with Missed Intervention.}
\label{fig:s2.h.m}
\end{figure}

\begin{figure}[h!]
\centering
\frame{\includegraphics[width=1.0\linewidth]{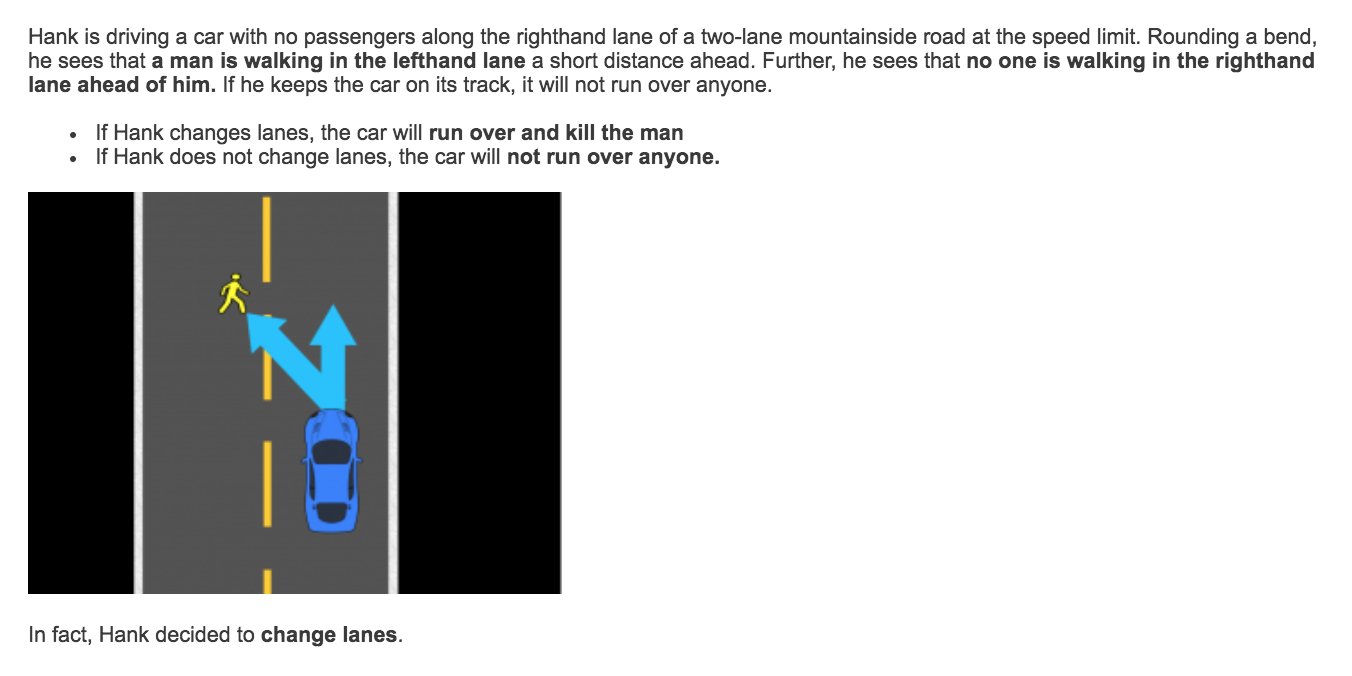}}
\caption{Vignette for Human only (Regular Car) with Bad Intervention.}
\label{fig:s2.h.b}
\end{figure}

\begin{figure}[h!]
\centering
\frame{\includegraphics[width=1.0\linewidth]{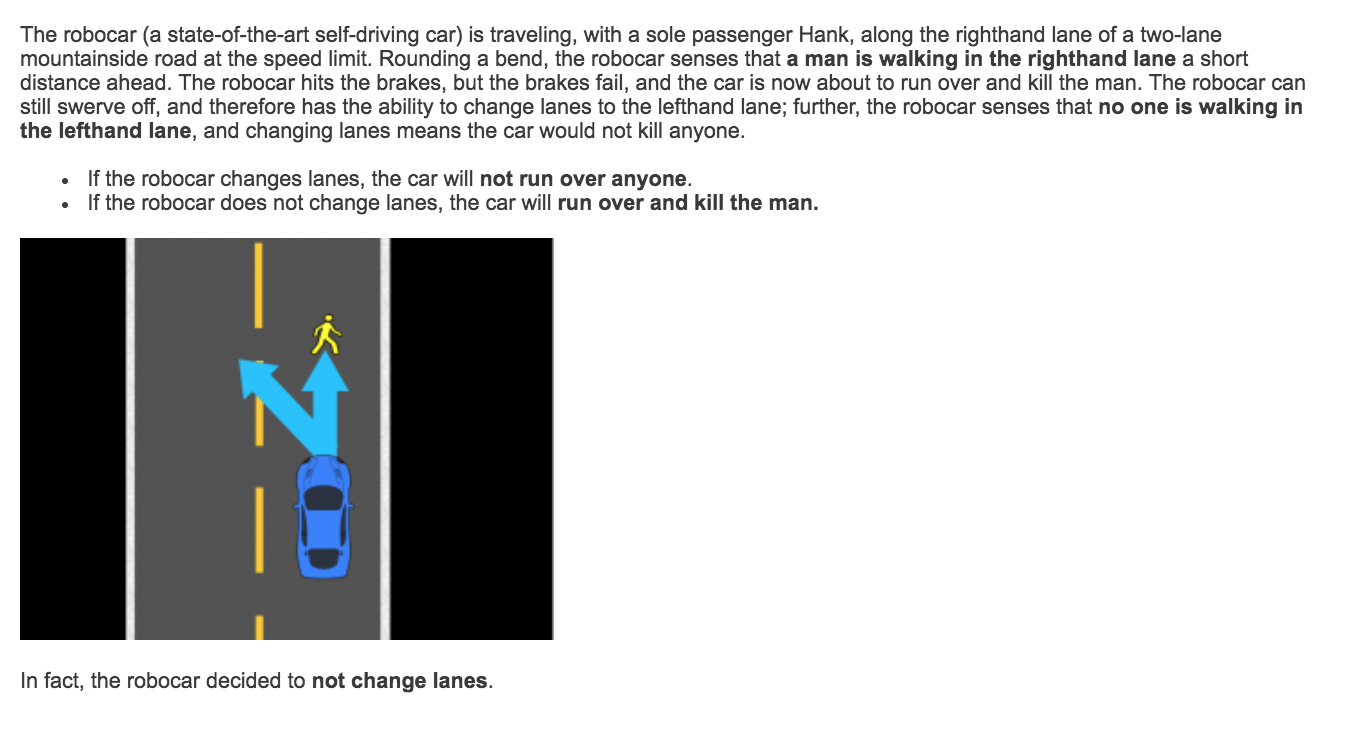}}
\caption{Vignette for Machine only (Fully Autonomous) with Missed Intervention.}
\label{fig:s2.m.m}
\end{figure}

\begin{figure}[h!]
\centering
\frame{\includegraphics[width=1.0\linewidth]{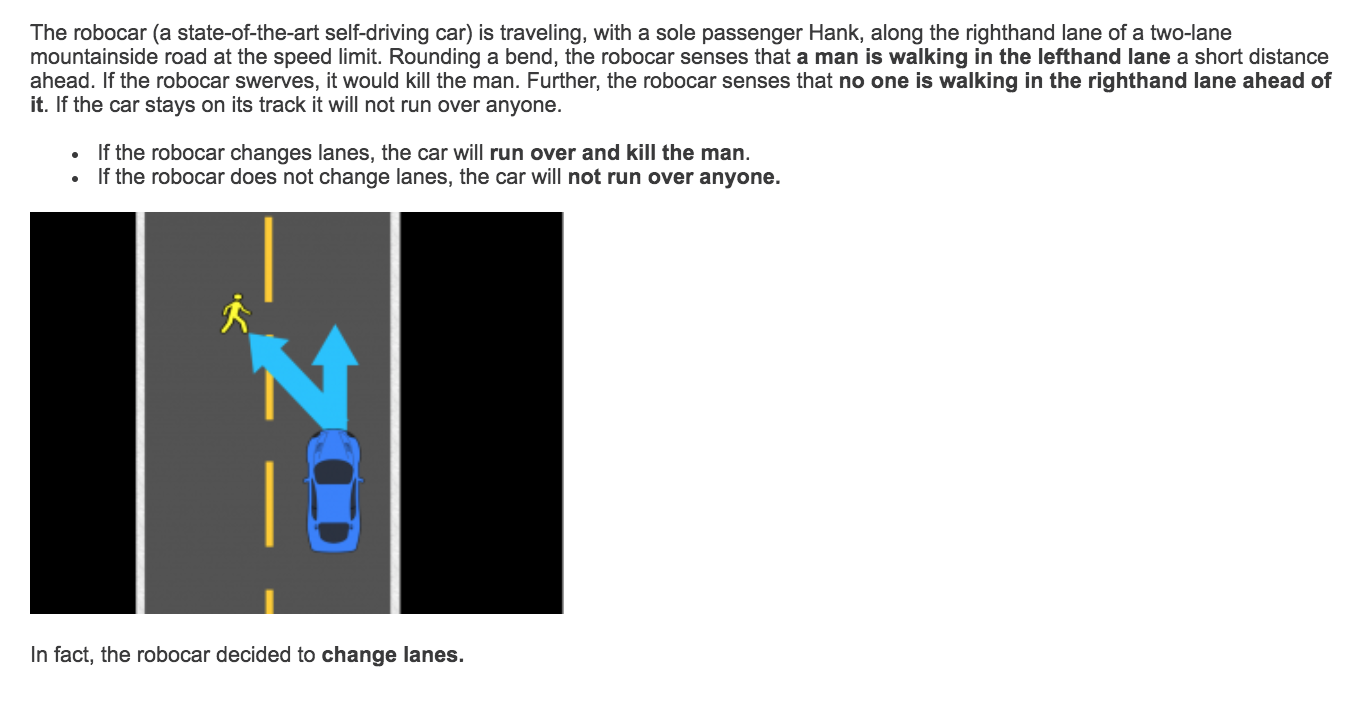}}
\caption{Vignette for Machine only (Fully Autonomous) with Bad Intervention.}
\label{fig:s2.m.b}
\end{figure}

\clearpage

\subsection*{Study 3}
\begin{figure}[h!]
\centering
\frame{\includegraphics[width=1.0\linewidth]{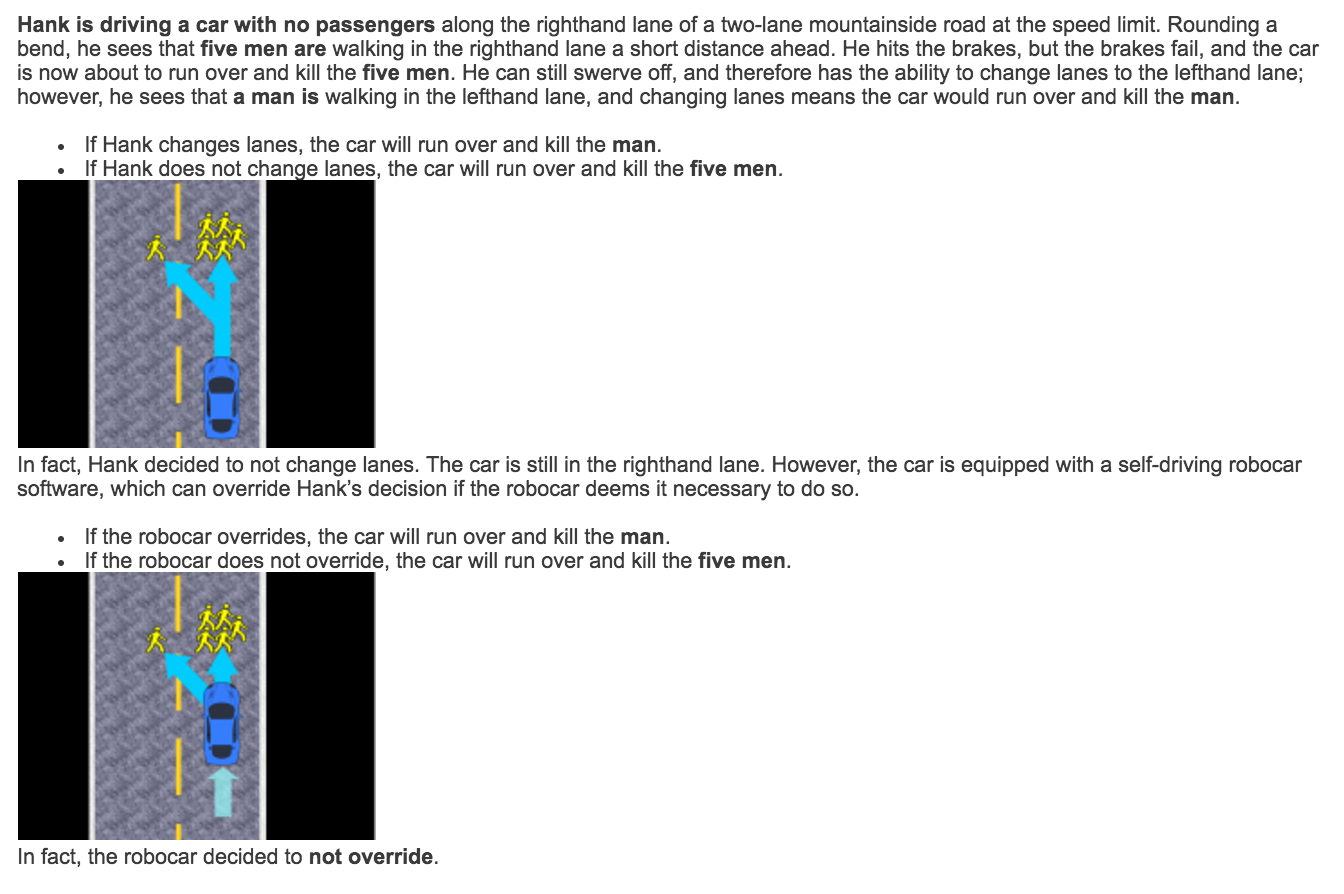}}
\caption{Vignette for Human-Machine with Missed Intervention.}
\label{fig:s3.hm.m}
\end{figure}

\begin{figure}[h!]
\centering
\frame{\includegraphics[width=1.0\linewidth]{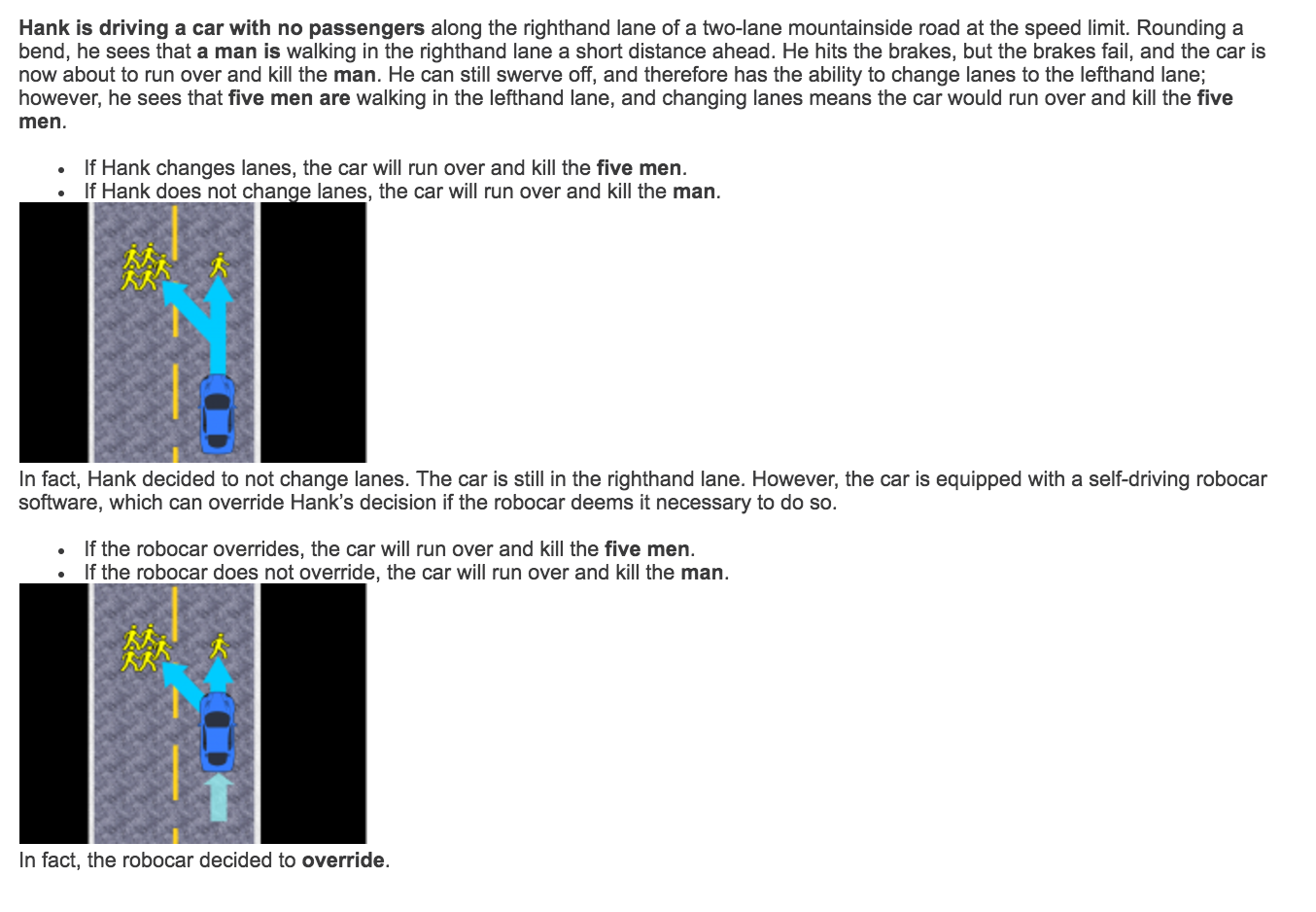}}
\caption{Vignette for Human-Machine with Bad Intervention.}
\label{fig:s3.hm.b}
\end{figure}

\begin{figure}[h!]
\centering
\frame{\includegraphics[width=1.0\linewidth]{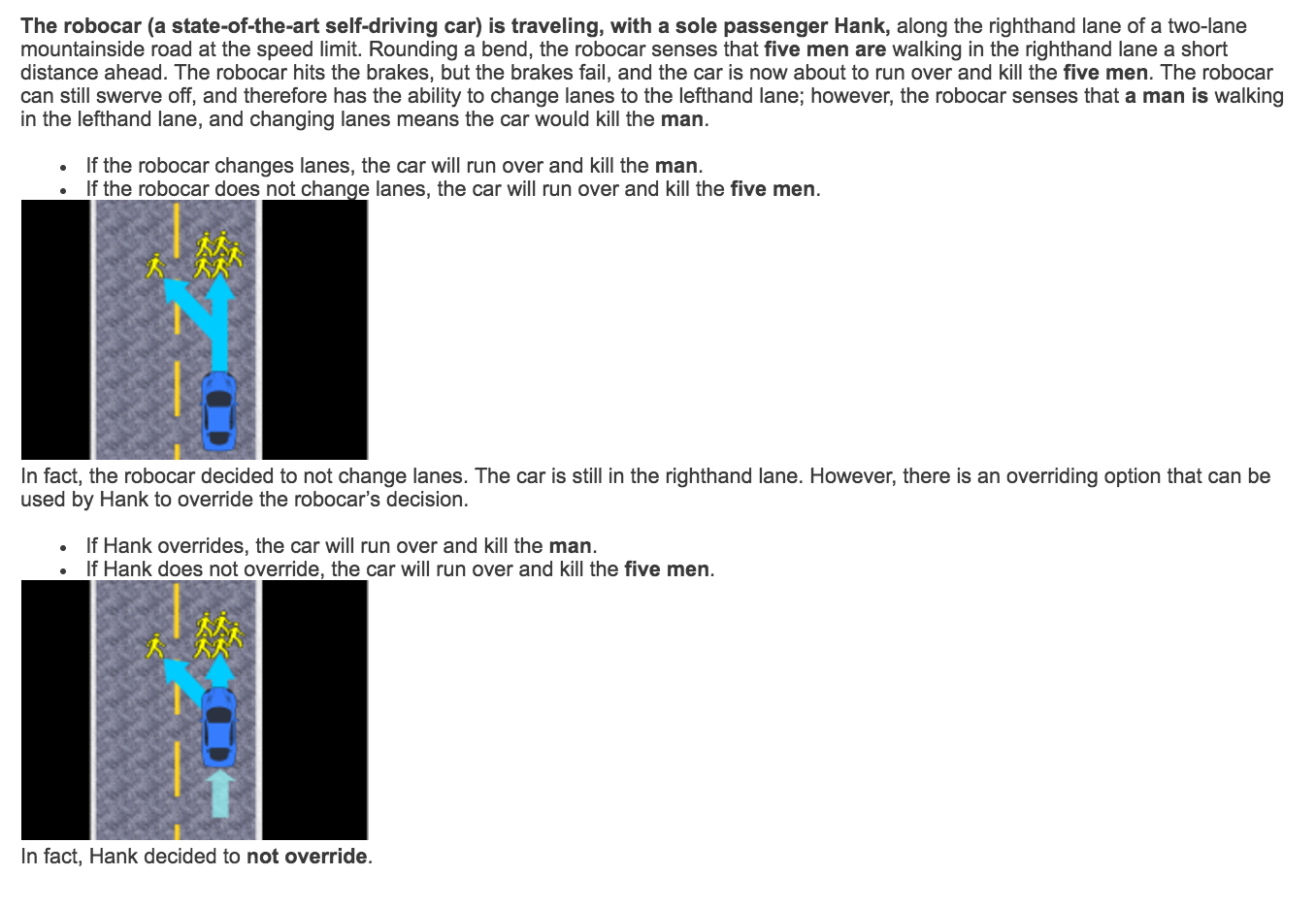}}
\caption{Vignette for Machine-Human with Missed Intervention.}
\label{fig:s3.mh.m}
\end{figure}

\begin{figure}[h!]
\centering
\frame{\includegraphics[width=1.0\linewidth]{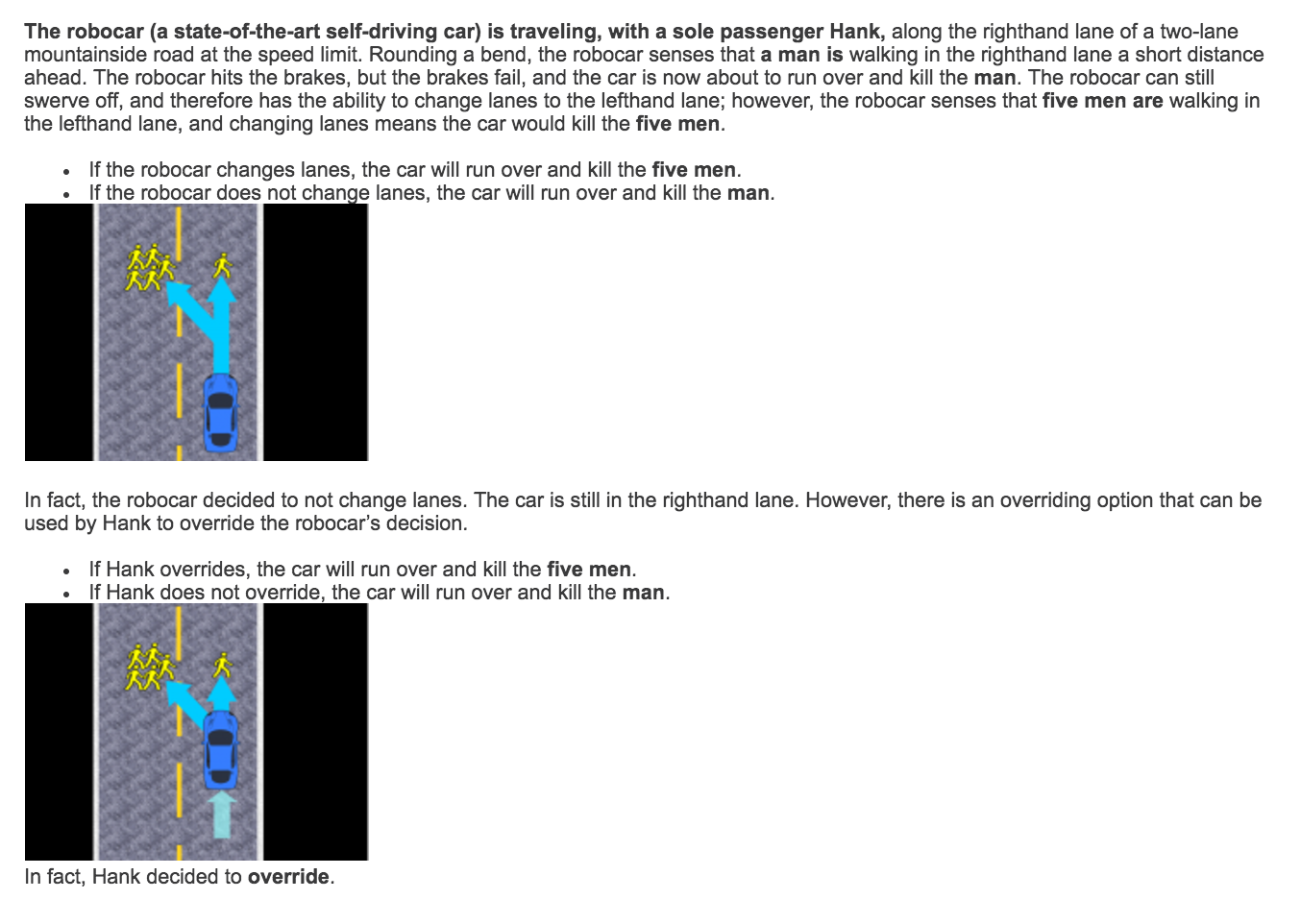}}
\caption{Vignette for Machine-Human with Bad Intervention.}
\label{fig:s3.mh.b}
\end{figure}

\begin{figure}[h!]
\centering
\frame{\includegraphics[width=1.0\linewidth]{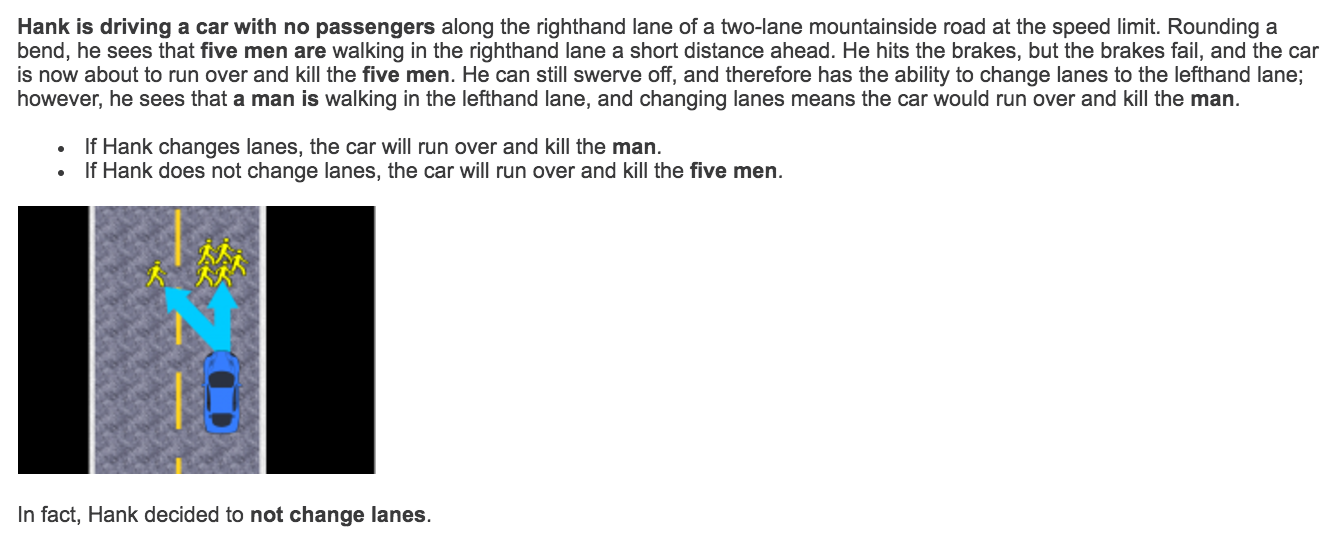}}
\caption{Vignette for Human only (Regular Car) with Missed Intervention.}
\label{fig:s3.h.m}
\end{figure}

\begin{figure}[h!]
\centering
\frame{\includegraphics[width=1.0\linewidth]{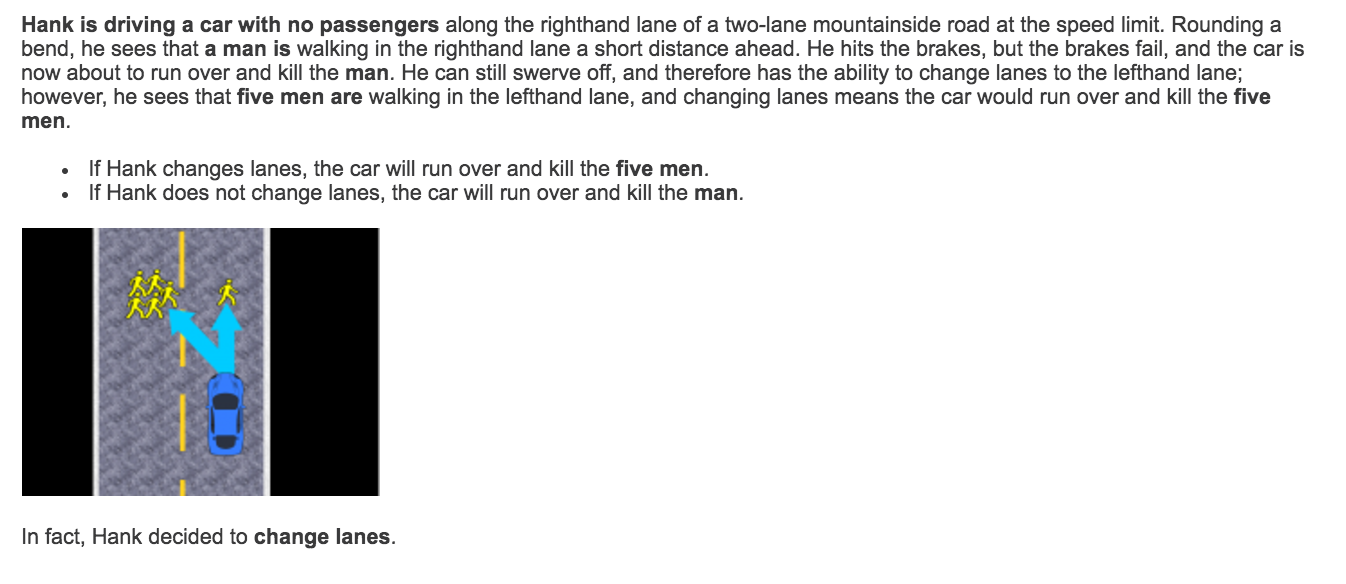}}
\caption{Vignette for Human only (Regular Car) with Bad Intervention.}
\label{fig:s3.h.b}
\end{figure}

\begin{figure}[h!]
\centering
\frame{\includegraphics[width=1.0\linewidth]{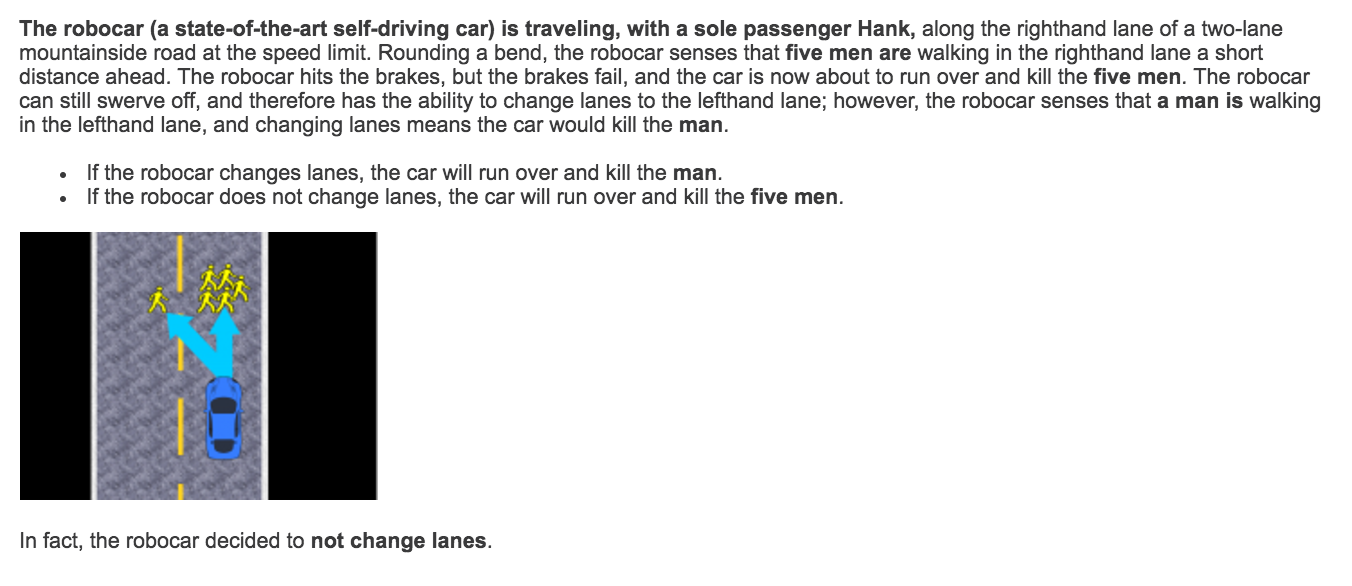}}
\caption{Vignette for Machine only (Fully Autonomous) with Missed Intervention.}
\label{fig:s3.m.m}
\end{figure}

\begin{figure}[h!]
\centering
\frame{\includegraphics[width=1.0\linewidth]{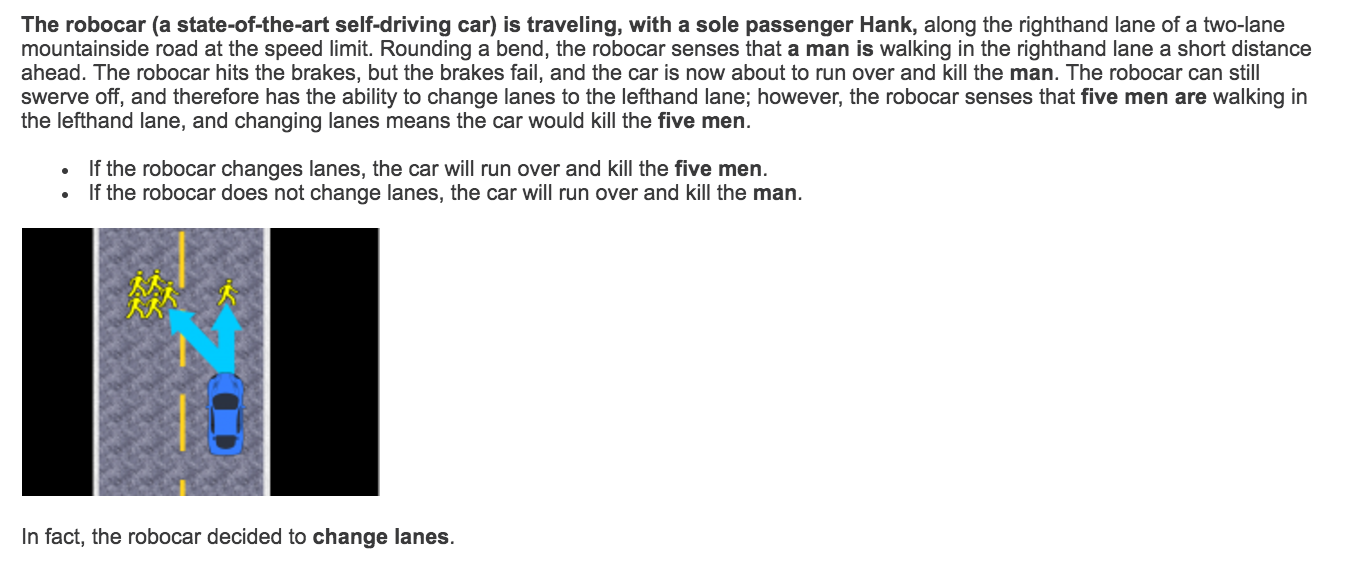}}
\caption{Vignette for Machine only (Fully Autonomous) with Bad Intervention.}
\label{fig:s3.m.b}
\end{figure}

\subsection*{Questions}

\begin{figure}[h!]
\centering
\frame{\includegraphics[width=1.0\linewidth]{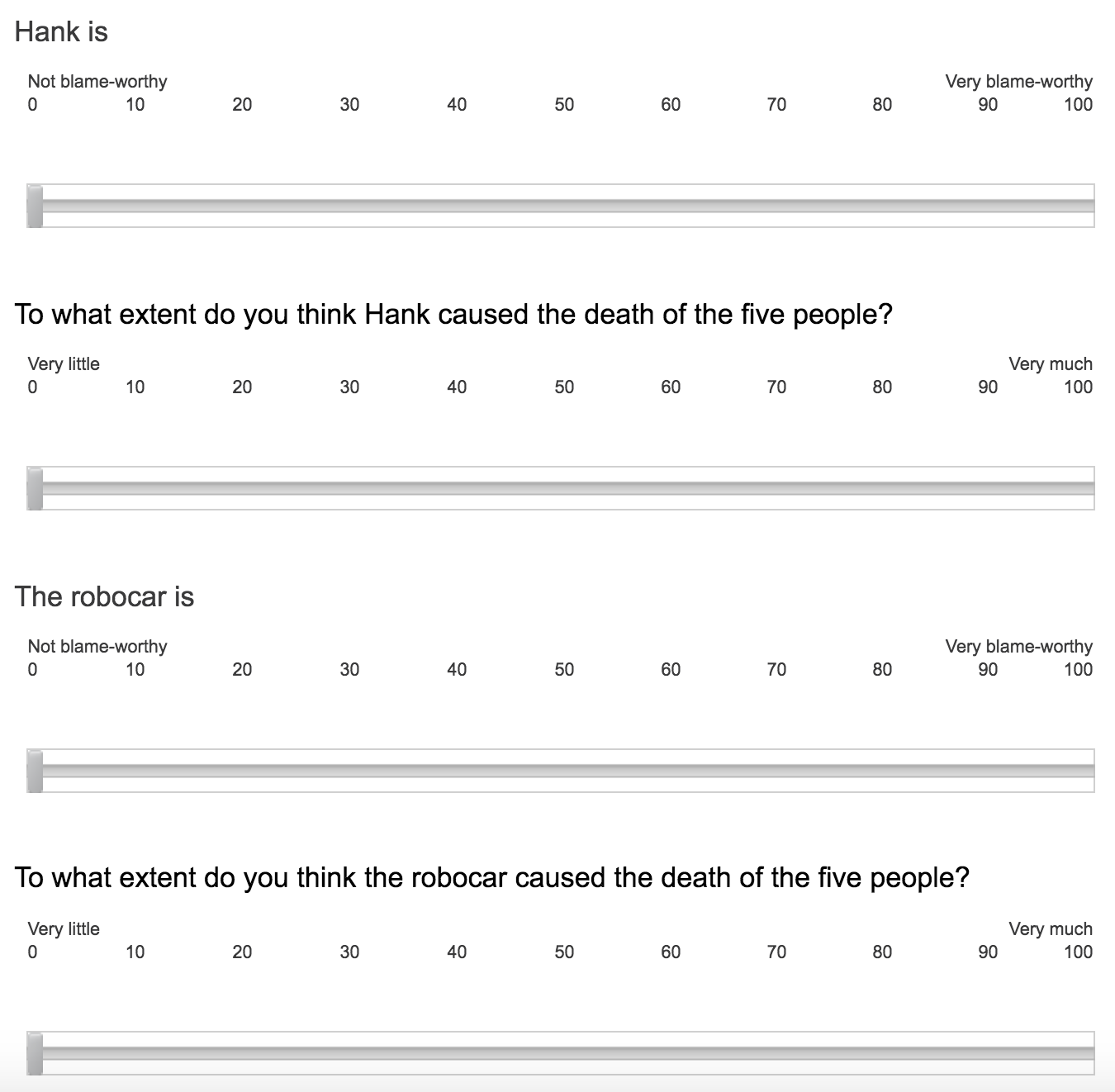}}
\caption{Questions asked for all studies and all cases, where Industry is the \emph{Robocar}. Robocar is replaced with robocar company or robocar programmer in other cases.}
\label{fig:qs}
\end{figure}

In all vignettes, after each scenario, four questions are asked: (Blame vs. Causal Responsibility) x (User vs. Industry). Industry is car in study 1; car or company in study 2; and car, company or programmer in study 3. See Figure \ref{fig:qs}.

\begin{figure}[h!]
\centering
\frame{\includegraphics[width=1.0\linewidth]{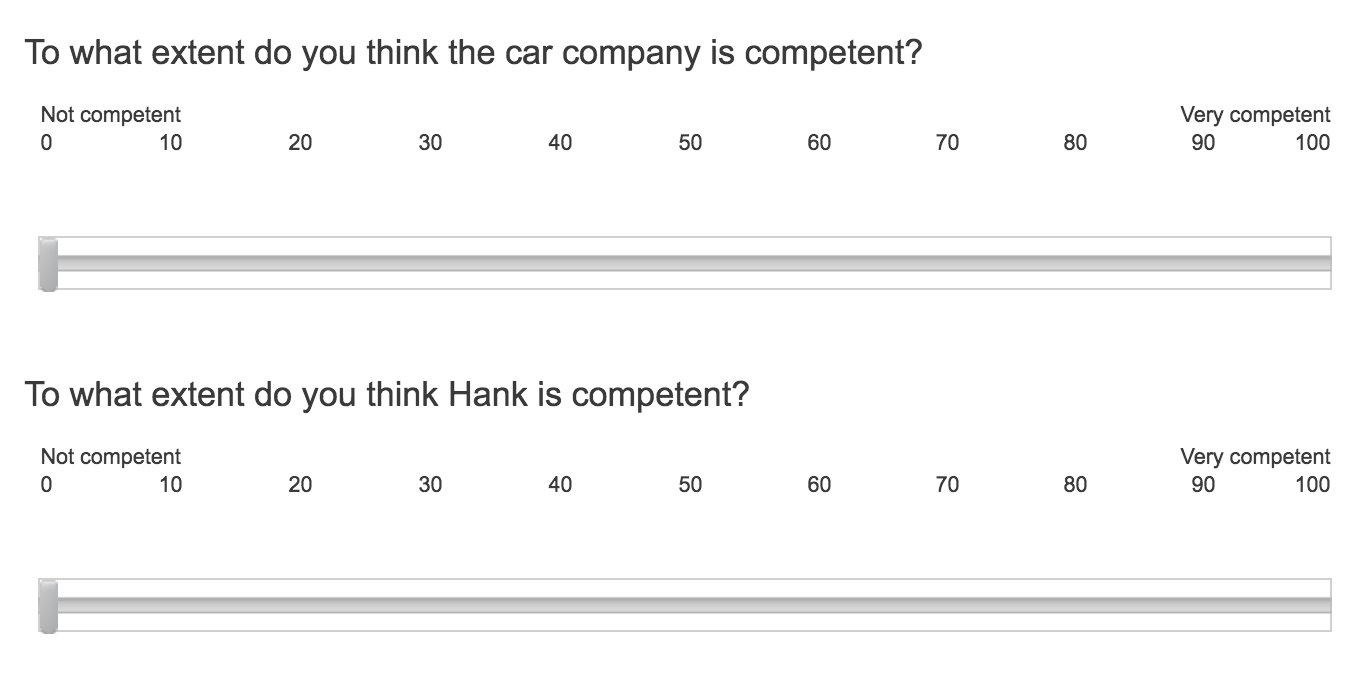}}
\caption{Competence questions asked in study 2 along with the other four questions, where Industry is the \emph{car company}.}
\label{fig:qs.comp}
\end{figure}

In study 2, in addition to the four questions above two more questions about competence are also asked (competence of Hank and competence of industry representative). See Figure \ref{fig:qs.comp}.

\begin{figure}[h!]
\centering
\frame{\includegraphics[width=1.0\linewidth]{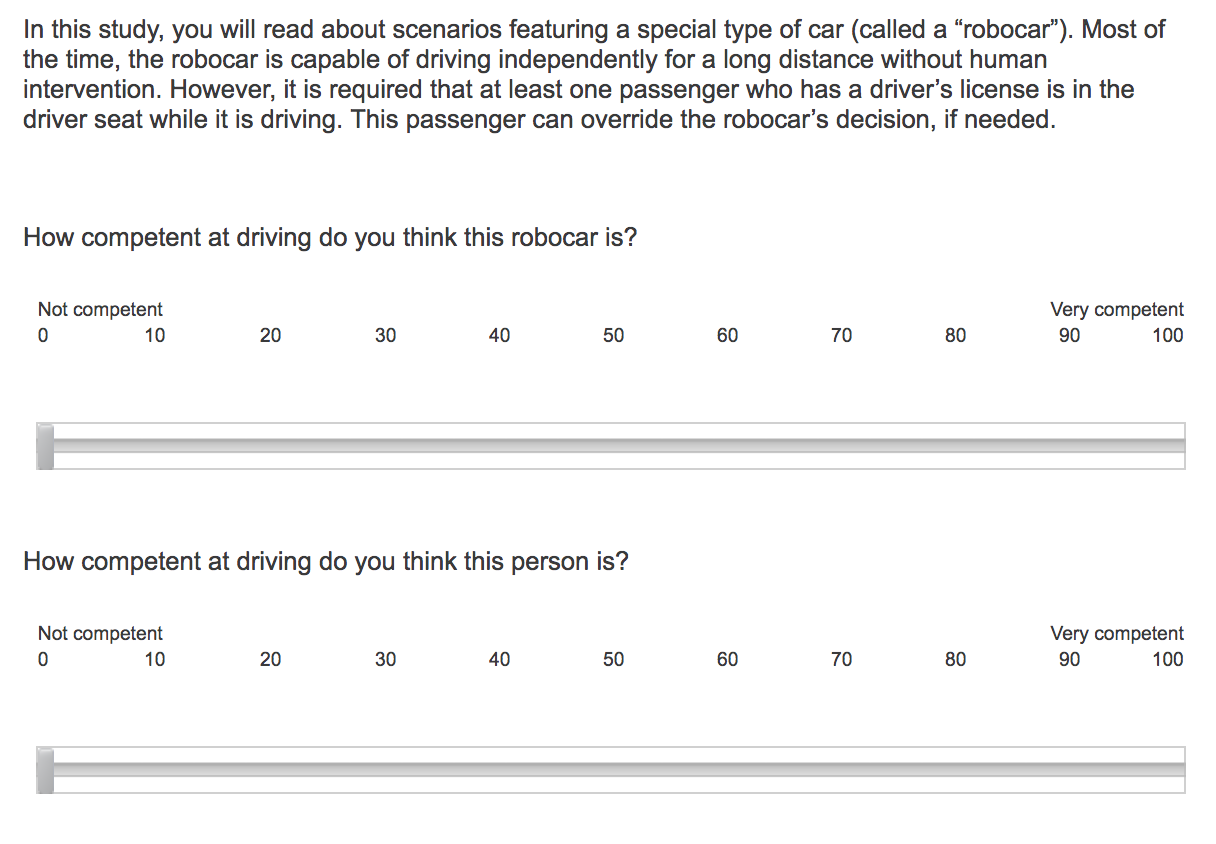}}
\caption{Description and Competence questions about Machine-Human from study 1.}
\label{fig:qs.s1.mh}
\end{figure}

\begin{figure}[h!]
\centering
\frame{\includegraphics[width=1.0\linewidth]{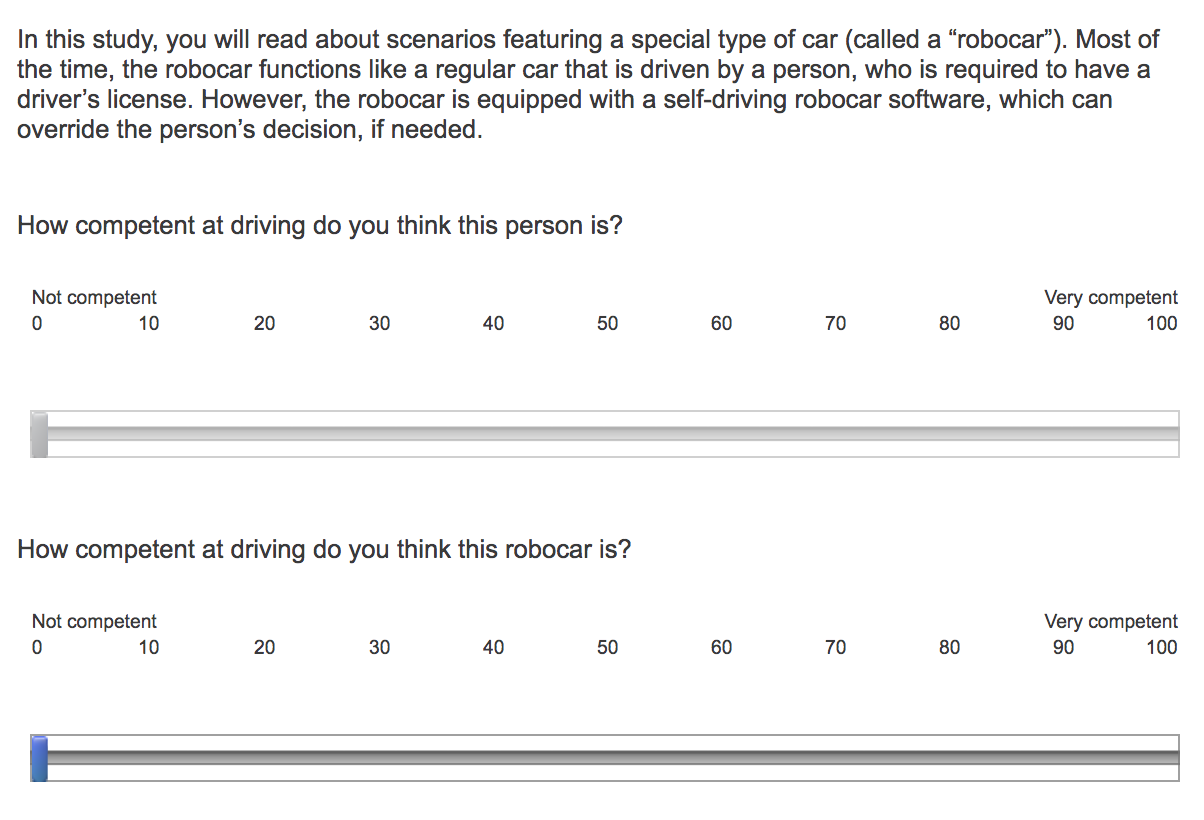}}
\caption{Description and Competence questions about Human-Machine from study 1.}
\label{fig:qs.s1.hm}
\end{figure}

\begin{figure}[h!]
\centering
\frame{\includegraphics[width=1.0\linewidth]{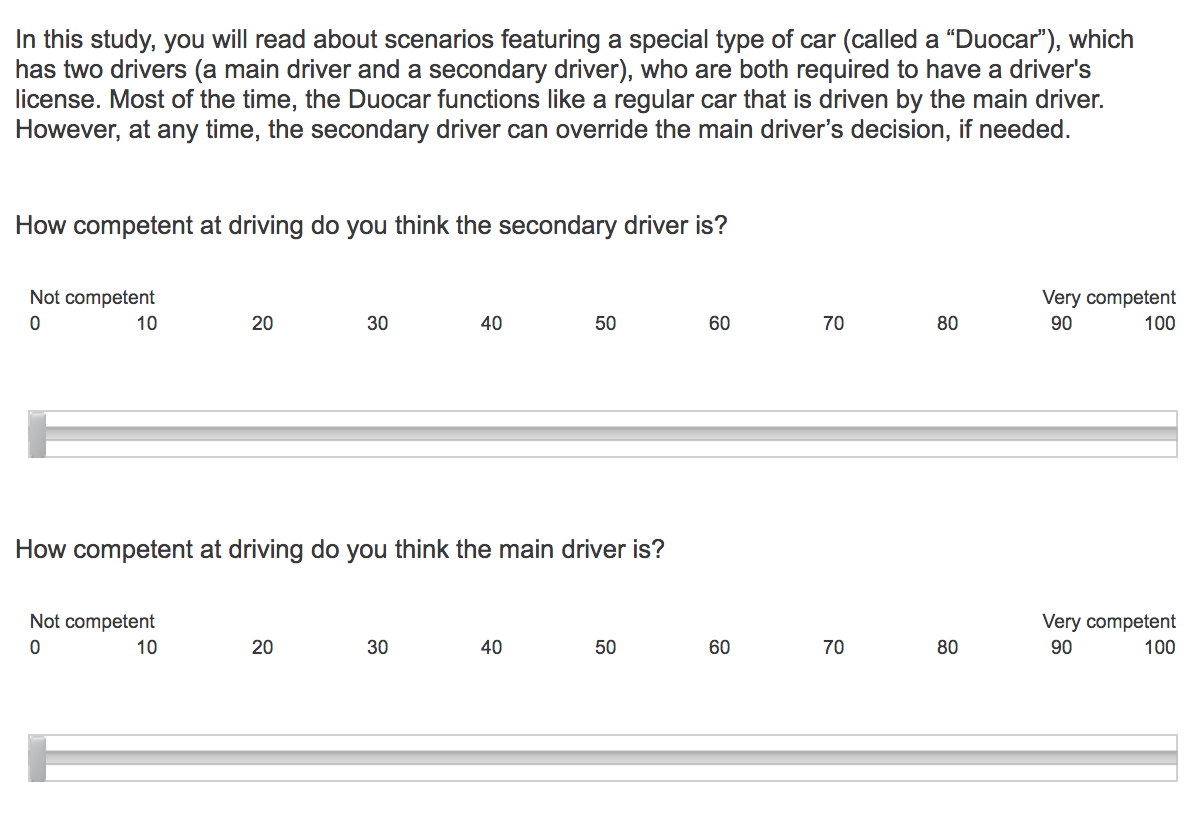}}
\caption{Description and Competence questions about Human-Human from study 1.}
\label{fig:qs.s1.hh}
\end{figure}

\begin{figure}[h!]
\centering
\frame{\includegraphics[width=1.0\linewidth]{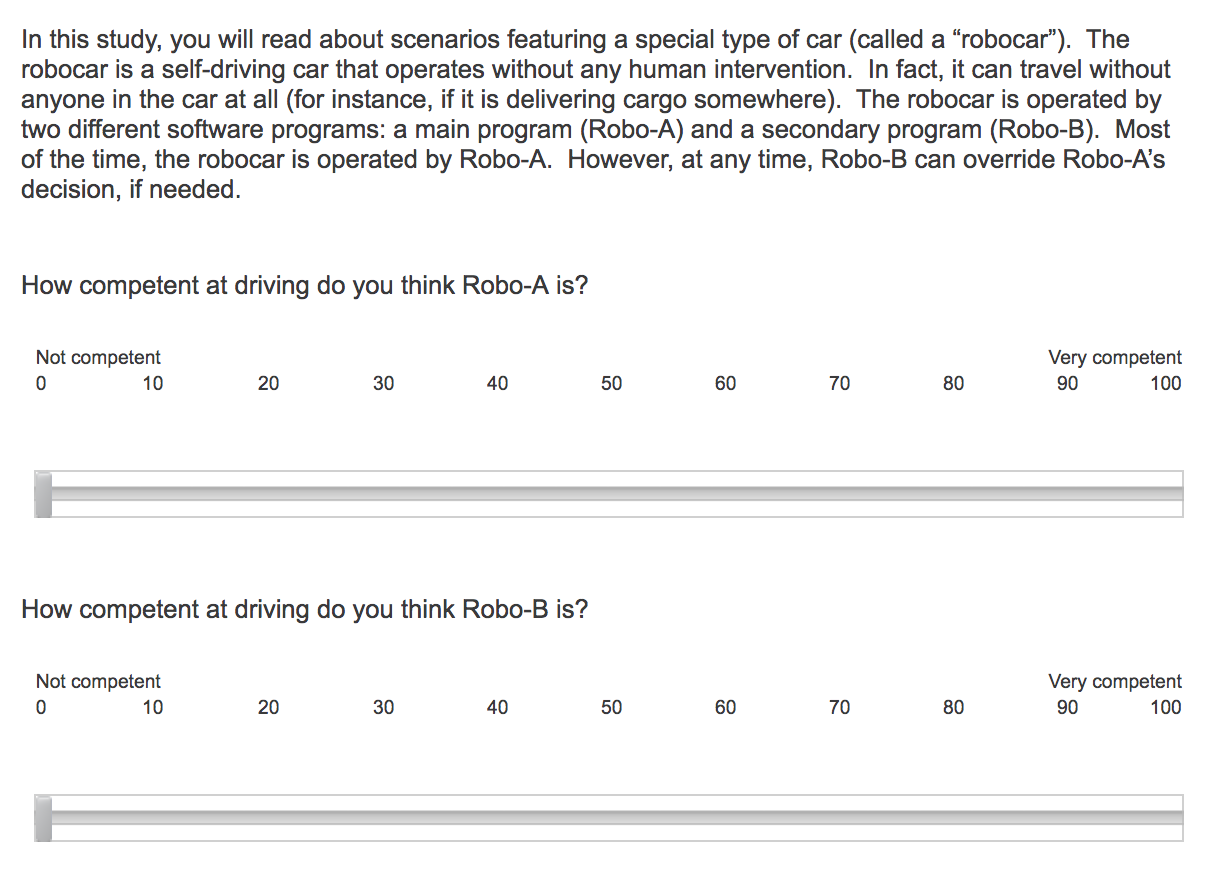}}
\caption{Description and Competence questions about Machine-Machine from study 1.}
\label{fig:qs.s1.mm}
\end{figure}

In study 1, competence questions are asked separately before respondents are presented with scenarios. In each condition, a description of the car regime is provided, and two questions about competence (User vs. Industry). See Figures \ref{fig:qs.s1.mh} -- \ref{fig:qs.s1.mm}.

\end{document}